\documentclass[pdflatex,sn-mathphys]{sn-jnl}

\jyear{2022}%

\theoremstyle{thmstyleone}%
\theoremstyle{thmstyletwo}%

\theoremstyle{thmstylethree}%
\newtheorem{definition}{Definition}%

\begin{document}

\title[ ]{A Survey of Automated Data Augmentation Algorithms for Deep Learning-based Image Classification Tasks}


\author*[1]{\fnm{Zihan} \sur{Yang}}\email{zihany1@student.unimelb.edu.au}

\author[1]{\fnm{Richard O.} \sur{Sinnott}}\email{rsinnott@unimelb.edu.au}

\author[1]{\fnm{James} \sur{Bailey}}\email{baileyj@unimelb.edu.au}

\author[1]{\fnm{Qiuhong} \sur{Ke}}\email{qiuhong.ke@unimelb.edu.au}

\affil*[1]{\orgdiv{Faculty of Engineering and Information Technology}, \orgname{The University of Melbourne}, \orgaddress{\street{700 Swanston Street}, \city{Melbourne}, \postcode{3010}, \state{Victoria}, \country{Australia}}}


\abstract{In recent years, one of the most popular techniques in the computer vision community has been the deep learning technique. As a data-driven technique, deep model requires enormous amounts of accurately labelled training data, which is often inaccessible in many real-world applications. A data-space solution is Data Augmentation (DA), that can artificially generate new images out of original samples. Image augmentation strategies can vary by dataset, as different data types might require different augmentations to facilitate model training. However, the design of DA policies has been largely decided by the human experts with domain knowledge, which is considered to be highly subjective and error-prone. To mitigate such problem, a novel direction is to automatically learn the image augmentation policies from the given dataset using Automated Data Augmentation (AutoDA) techniques. The goal of AutoDA models is to find the optimal DA policies that can maximize the model performance gains. This survey discusses the underlying reasons of the emergence of AutoDA technology from the perspective of image classification. We identify three key components of a standard AutoDA model: a search space, a search algorithm and an evaluation function. Based on their architecture, we provide a systematic taxonomy of existing image AutoDA approaches. This paper presents the major works in AutoDA field, discussing their pros and cons, and proposing several potential directions for future improvements.}




\maketitle

\section{Introduction}\label{Sec: intro}

Promoted by recent advances in neural network architectures, deep learning has made great progress in Computer Vision (CV) \cite{krizhevsky2012imagenet, szegedy2015going, simonyan2014very, he2016deep}. In particular, deep learning models have been successfully applied to image classification tasks in diverse areas from medical imaging \cite{esteva2017dermatologist, shin2016deep} to agriculture \cite{zheng2019cropdeep, kamilaris2018deep}. However, to achieve enhanced performance, deep learning, as a data-driven technology, places significant demands on both the quantity and quality of data for model training and testing. Effectively training a supervised model highly relies on enormous amounts of annotated data, which is often challenging for most practical applications \cite{shijie2017research}. 

To address the issue of data insufficiency, Data Augmentation (DA) is widely utilized. In general, data augmentation refers to the process of artificially generating data samples to increase the size of training data \cite{lemley2017smart}. In the imaging domain, this is usually done by applying image Transformation Functions (TFs), such as translation, rotation or flipping. For computer vision tasks, image DA has been utilised in nearly all supervised neural network architectures to increase data volume and variety, including traditional data-driven models \cite{cirecsan2010deep, dosovitskiy2015discriminative, graham2014fractional, sajjadi2016regularization}, and few/zero-shot learning \cite{rios2018few}. Besides supervised approaches, DA techniques are also extensively applied in the field of unsupervised learning. For example, contrastive self-supervised learning relies on image transformations to incorporate data invariance in the representation space across various augmentations \cite{bachman2019learning}. 

In the context of image augmentation, a DA policy refers to a set of image operations, which are used to transform the image data. When applying image DA, choosing a carefully designed augmentation scheme (i.e. DA policy) is necessary to improve the effectiveness of DA and hence the associated network training \cite{krizhevsky2012imagenet, paschali2019data}. For instance, data augmented by random image operations can be redundant. But overly aggressive TFs might corrupt the original semantics, and introduce potential biases into the training dataset \cite{graham2014fractional}. Therefore, different datasets or domains may require different types of augmentations. Specifically, when standard supervised approaches are applied, classification tasks with limited data may require label-preserving augmentations to provide direct semantic supervision. However, for few/zero-shot learning models, more emphasis is placed on increasing data diversity in order to generate an enriched training set \cite{wang2020generalizing}, which might promote more aggressive augmentation TFs. 

In spite of the ubiquity and importance of DA techniques, there is little selection strategy in DA policy design when given certain datasets. Unlike other machine learning topics that have been thoroughly explored, less attention has been put into finding effective DA policies to benefit particular dataset, and hence improve the model accuracy. Instead, DA policies are often intuitively decided based on past experience or limited trials \cite{dao2019kernel}. Decisions on augmentation strategies are still made by human experts, based on prior knowledge. For example, the standard augmentation policy on ImageNet data was proposed in 2012 \cite{krizhevsky2012imagenet}. This is still widely used in most contemporary networks without much modification \cite{cubuk2019autoaugment}. Furthermore, criteria for selecting good augmentation methods on different datasets may greatly vary due to the nature of given tasks. The traditional trial-and-error approach based on training loss or accuracy can give rise to extensive, redundant data collections, wasting computational efforts and resources. 

Motivated by progress in Automated Machine Learning (AutoML), there has been a rising interest in automatically searching effective augmentation policies from training data \cite{cubuk2019autoaugment, lim2019fast, hataya2020faster, ho2019population}. Such technique are often referred to as Automated Data Augmentation (AutoDA). Recent research has found that instead of manually design the DA schemes, directly learning a DA strategy from the target dataset has the potential to significantly improve model performance \cite{lemley2017smart, tran2017bayesian, devries2017dataset, zoph2020learning}. Specifically, the DA policy that can yield the most performance gain on classification model is regarded as the optimal augmentation policy for a given dataset. Among various contemporary works, AutoAugment (AA) stands out as the first AutoDA model, achieving state-of-the-art results on several popular image classification datasets, including CIFAR-10/100 \cite{krizhevsky2009learning}, ImageNet \cite{deng2009imagenet} and SVHN \cite{netzer2011reading}. More importantly, AA provides essential theoretical foundation for later works that support automated augmentation \cite{lim2019fast, ho2019population, hataya2020faster, tian2020improving, cubuk2020randaugment}. 

The progress of automating DA policy search can potentially change the existing process of model training. AutoDA model can automatically select the most effective combination of augmentation transformations to form the final DA policy. Once the optimal augmentation policy is found, the training set augmented by the learned policy can dramatically boost the model performance without extra input. Furthermore, AutoDA methods can be designed to be directly applied on the datasets of interest. The optimal DA policy learned from the data is regarded as the best augmentation formula for the target task, and hence it should guarantee the best model performance. Another desirable aspect of AutoDA techniques is their transferability. According to the findings in \cite{cubuk2019autoaugment}, learned DA policies can also be applied on other similar datasets with promising results.

Although considerable progress has been made for DA policy search, there is still a lack of comprehensive survey that can systematically summarize the diverse methods. To the best of our knowledge, no one has conducted a qualitative comparison of existing AutoDA methods or provided a systematic evaluation of their advantages and disadvantages. To fill this gap, this paper aims to identify the current state of research in the field of Automated Data Augmentation (AutoDA), especially for image classification tasks.

In this paper, we mainly review contemporary AutoDA works in imaging domain. We provide a systematic analysis, identifying three key components of standard AutoDA techniques, i.e. search space, search algorithm and evaluation function. Based on the different choices on search algorithms in reviewed works, we then propose a two-layer taxonomy of all AutoDA approaches. We also evaluate AutoDA approaches in terms of the efficiency of search algorithm, as well as the final performance of trained classification model. Through comparative analysis, we identify major contributions and limitations of these methods. Lastly, we summarise several main challenges and propose potential future directions in AutoDA field. 

Our main contributions can be summarized as follows:

\begin{enumerate}[1.]
    \item background on image data augmentation, including traditional approaches and Automated Data Augmentation (AutoDA) models (Section \ref{Sec: background});
    \item introduction of three key components within standard AutoDA models, along with evaluation metrics and benchmarks used in most works (Section \ref{Sec: autoda});
    \item a hierarchical taxonomy of the mainstream AutoDA algorithms for image classification tasks from the perspective of hyper-parameter optimization (Sections \ref{Sec: taxonomy});
    \item thorough review of each AutoDA method in the taxonomy, detailing the search algorithm applied (Section \ref{Sec: two} and Section \ref{Sec: one});
    \item discussion about the current state of AutoDA technique, as well as the existing challenges and potential opportunities in future (Section \ref{Sec: discuss}).
\end{enumerate}

\section{Background}
\label{Sec: background}

This section introduces background information about data augmentation in the computer vision field with focus on image classification tasks. We first provide a general overview of how DA technique developed and been applied to computer vision tasks. Then we briefly describe several traditional image processing operations that are involved in most AutoDA models. Finally, we discuss the recent advances in AutoDA techniques and how such techniques relate to Automated Machine Learning (AutoML).

\subsection{Historical Overview of Image Data Augmentation}

The early application of image augmentation can be traced back to LeNet-5 \cite{lecun1998gradient}, where a Data Augmentation (DA) technique was applied by distorting images for recognizing handwritten and machine-printed characters. This work was one of the earliest pre-trained Convolutional Neural Networks (CNNs) that used DA for image classification tasks. Generally, DA can be regarded as an oversampling method. The objective of oversampling is to mitigate the negative influence of limited data or class imbalances by increasing data samples. A naive approach for oversampling is random oversampling, which randomly duplicates data points in minority classes until a desired data amount or data distribution is achieved. However, the duplicate images created by this technique can result in model overfitting towards the minority class. This problem becomes even more notable when deep learning technique is used. To add more variety to generated samples, DA via image transformations has emerged.  

The most early famous use case of image DA was AlexNet model \cite{krizhevsky2012imagenet}. AlexNet significantly improved classification results on ImageNet data \cite{deng2009imagenet} through the use of a revolutionary CNN architecture. In their work, image augmentation was used to artificially expand the dataset. Multiple image operations were applied to the original training set, including random cropping, horizontal flipping and colour adjustment in RGB space. These transformation functions helped mitigate overfitting problems during model training. According to the experimental results in \cite{krizhevsky2012imagenet}, image DA reduced the final error rate by approximately $1\%$. Since then, image augmentation has been regarded as a necessary pre-processing procedure before training complex CNNs, from VGG \cite{simonyan2014very} to ResNet \cite{he2016deep} and Inception \cite{szegedy2016rethinking}. 

Image augmentation is not limited to the basic image processing. Following the proposal of Generative Adversarial Network (GAN) in \cite{goodfellow2014generative}, related works flourished in the following decade. Among them, the most influential technique was Neural Architecture Search (NAS) \cite{zoph2016neural}. NAS is a type of AutoML technique, which is the process of searching for model architectures through automation. The advancement of NAS greatly promoted the development of DA technology in the imaging field. Applying concepts and techniques from NAS and AutoML has gained increasing interest in the CV community. Recent progress include Neural Augmentation \cite{perez2017effectiveness}, which tests the effectiveness of GANs in image augmentation; Smart Augmentation \cite{lemley2017smart}, which generates synthetic image data using neural networks; and AA \cite{cubuk2019autoaugment}, which is aimed at the automation of image transformation selection for DA. The latter work forms the basis for AutoDA and is the focus of this survey. 

Most of the augmentation methods mentioned before were designed for image classification. The ultimate goal of image DA in classification tasks is to improve the predication accuracy of discriminative models. However, the same technique is applicable for other computer vision tasks, for example Object Detection (OD), where image augmentation can be combined with advanced deep neural networks including YOLO \cite{redmon2016you} and R-CNN series \cite{girshick2014rich, girshick2015fast, ren2015faster}. Semantic segmentation task \cite{long2015fully} can also benefit from DA before training complex networks such as U-Net \cite{ronneberger2015u}. In this study, we particularly focus on the application of Automated DA (AutoDA) for image classification tasks, as there exists more published datasets in this domain that allow to conduct a fair evaluation. For some AutoDA methods, we also discuss the possibilities of applying them in object detection tasks if there are experimental results available.

\subsection{Traditional Image Augmentation Techniques}

Image augmentation aims to enhance both the quantity and quality of datasets so that neural networks can be better trained \cite{shorten2019survey}. Usually, DA does this in two ways, either through traditional image operations or based on deep learning technology. Traditional augmentation often emphasizes on preserving the image's original label and transforms existing images into a new form \cite{shorten2019survey}. This method can be achieved through various image processing approaches, including but not limited to geometric transformations, adjustment in colour space or even combinations of them. 

Another augmentation technique based on deep learning attempts to generate synthetic images as the training set. Major techniques involve Mixup augmentation \cite{zhang2017mixup}, GANs and transformations in feature space \cite{devries2017dataset}. Due to the complexity of deep learning DA, only the basic image processing operations are considered in recent automated DA methods. Hence, we focus on the basic image transformations that can be easily parameterized. The rest of this section briefly introduces several basic image processing functions that are usually considered in AutoDA models, including geometric transformations, flipping, rotation, cropping, colour adjustment and kernel filters. Another two augmentation algorithms are also covered due to their presence in AutoAugment work \cite{cubuk2019autoaugment}, namely Cutout \cite{devries2017improved} and SamplePairing \cite{inoue2018data}.

\subsubsection{Geometric Transformations}

The simplest place to start image augmentation is using geometric transformation functions, such as image translation or scaling. These operations are easy to implement and can also be combined with other transformations to form more advanced DA algorithms. One important thing when applying such operations is whether they can preserve the original image label after the transformation \cite{bagherinezhad2018label}. From the perspective of image augmentation, the ability to keep label consistency can also be called safety feature of transformation functions \cite{shorten2019survey}. In other words, transformations that may risk corrupting annotation information are considered to be unsafe. In general, geometric transformations tend to preserve the labels as they only change the position of key features. However, depending on the magnitude of the transformation function, the application of the chosen operation might not always be safe. For example, translation of the $y$ axis with a high magnitude may end up completely shifting the object-of-interest outside of the visible area, therefore it fails to preserve the label of the post-processed image. 

\subsubsection{Rotation}

Rotating the image by a given angle is another common DA technique. It is a special type of geometric transformation, which also has the risk of removing meaningful semantic information from the visible area. Aggressive operations with a large rotation angle are usually unsafe, especially for text-related data, e.g. "6" and "9" in SVHN data \cite{netzer2011reading}. However, according to \cite{shorten2019survey}, slight rotation within the range of $1^{\circ}$ to $30^{\circ}$ can be helpful for most image classification tasks. 

\subsubsection{Flipping}

Flipping is different from rotation augmentation as it generates mirror images. Flipping can be done either horizontally or vertically, while the former is more commonly used in computer vision \cite{shorten2019survey}. This is one of the simple yet most effective augmentation techniques on image data, especially for CIFAR-10/100 \cite{krizhevsky2009learning} and ImageNet \cite{deng2009imagenet}. The safety feature of flipping largely depends on the type of input data. For normal classification or object detection tasks, flipping augmentation preserves the original label. But it can be unsafe for data involving digits or texts, such as SVHN data \cite{netzer2011reading}. 

\subsubsection{Cropping}

Cropping is not only a basic DA method, but also an important pre-processing step before training when there are various sizes of image samples in the input data. Before being fed into the model, training data needs to be cropped into a unified $x \times y$ dimension for later matrix calculations. As an augmentation technique, cropping has a similar effect to geometric translation. Both augmentation methods remove part of the original image patch, while image translation keeps the same spatial resolution of the input and output image. In contrast, cropping will reduce the size of processed image. As described previously, cropping can be a safe/unsafe depends on its associated magnitude value. Aggressive operations might crop the distinguishable features, affecting label consistency, whereas a reasonable magnitude value helps to improve the quality of the training data.

\subsubsection{Colour Adjustment}

Adjusting values in colour space is another practical augmentation strategy that has been commonly adopted. Through the value jitters of single colour channels, it is possible to quickly obtain different colour representations of an image. These RGB values can also be manipulated through matrix operations to mimic different lighting conditions. Alternatively, self-defined rules on pixel values can be applied to implement transformations such as Solarize, Equalize, Posterize functions using in Python Image Library (PIL) \cite{umesh2012image}. Different from previous DA transformations, colour adjustment preserves the original size and content of input images. However, it might discard some colour information and thus might raise safety issues. For example, if colour is a discrimitative feature of an object of interest, when manipulating the colour values, the distinctive colour of the object may be hard to observe and hence confuse the model. The magnitude of colour transformation is again the determining factor that affects its safety property.

\subsubsection{Kernel Filters}

Instead of directly changing pixel values in colour space, they can be manipulated via kernel filters. This is a widely used technique in computer vision field for image processing. A filter is usually a matrix of self-defined numbers, with much smaller size than the input image. Depending on the element values in the matrix, kernel filters can provide various functionalities. The most common kernel filters include blurring and sharpening. To apply the kernel filter on on input image, we treat it as a sliding window, and scroll it over the whole image to get the pixel values out of matrix multiplications as our final output. A Gaussian kernel can cause blurring effect on the filtered image, which can better prepare the model for low-quality images with limited resolution. In contrast, a sharpening filter emphasizes the details in the image, which can help the model gain more information about the key features. 

\subsubsection{Cutout}

Besides simple transformations, another interesting augmentation technique is Cutout \cite{devries2017improved}. Cutout is inspired by the concept of dropout regularisation, performed on input data instead of embedded units within neural network \cite{zhong2020random}. This algorithm is specifically designed for object occlusion problems. Occlusion happens when some parts of the object are vague or occluded (hidden) by other non-relevant objects, in which case, only partial observation of the object is possible. This is a common problem especially in real-world scenarios. Cutout augmentation combats this by randomly cropping a small patch out of the original image to simulate the occlusion cases. Training on such transformed data, models are forced to learn from the whole picture rather than just a section of it, which enhances its ability to distinguish object features. Another convenient feature of the Cutout algorithm is that it can be applied along with other image augmentation methods, such as geometric or colour transformation, to generate more diverse training data. 

\subsubsection{SamplePairing}

SamplePairing \cite{inoue2018data} is an example of a complex augmentation algorithm that combines several simple transformations. It creates a completely new image by randomly choosing two data samples from the training set and mixing them. In standard SamplePairing, such combination is done by calculating the average of pixel values in two samples. The label of the generated images follows the first image and ignores the annotation of the second sample in the input pair. One of the advantages of SamplePairing augmentation is that it can create up to $N^2$ new data points out of dataset of size $N$ via simple permutation. SamplePairing is straightforward augmentation method that generates synthetic data points out of the original data. The enhancement of data quantity and variety significantly improves model performance and avoids model overfitting problems. This technique is especially helpful for computer vision tasks with limited training data. 

\subsection{Development of Automated Data Augmentation (AutoDA)}

With various image augmentation operations available, the question is how to choose an effective DA policy from these transformations for CV tasks. A naive solution is to apply random augmentations, generating vast amounts of transformed data for training. However, without appropriate control on the type and magnitude of augmentation TFs, the augmented data points might be simple duplicates or even semantically corrupted, which can lead to performance loss. Furthermore, overly augmented data might require excessive computational resources during model training, causing efficiency issues. A systematic selection strategy for a DA policy is therefore needed. A DA policy refers to a composition of various image distortion functions, which can be applied to training data for data augmentation. 

Despite extensive research on DA transformations, the selection of a given augmentation policy usually relies on human experts. Especially in the context of CV tasks, the decision on which image operations to use is mainly made by machine learning engineers based on past experience or domain knowledge. Therefore, the optimal strength of a given DA policy is highly task-specific. For example, geometric and colour transformations are commonly used in standard classification datasets, including CIFAR-10/100 \cite{krizhevsky2009learning} and ImageNet \cite{deng2009imagenet}. While resizing and elastic deformations are more popular on digit images such as MNIST \cite{sato2015apac, simard2003best} and SVHN \cite{netzer2011reading} datasets. There is no universal agreement on augmentation strategies for all types of CV tasks. In most cases, DA policies need to be manually selected based on prior knowledge. However, human effort involved in deep learning is usually considered biased and error-prone. There is no theoretical evidence to support the optimal human-decided DA policies. It is infeasible to manually search for the optimal DA policy that can achieve the best model performance. Additionally, without the help of advanced ML technique, finding an effective DA policy must rely on empirical results from multiple experiments, which can be excessively time-consuming. 

To reduce the potential bias and accelerate the design process, there has been increasing interest in automating the selection of DA policies. This technique is known as Automated Data Augmentation (AutoDA). The development of AutoDA is motivated by the advancements in Neural Architecture Search (NAS) \cite{zoph2016neural}, which automatically searches for the optimal architecture for deep neural networks instead of by manual approach. The majority of AutoDA techniques rely on different search algorithms to search for the most effective (optimal) augmentation policy for a given dataset. In the context of AutoDA, an optimal DA policy is the augmentation scheme that can yield the most performance gain and highest accuracy score. 

The earliest AutoDA work can be traced back to Transformation Adversarial Networks for Data Augmentations (TANDA) \cite{ratner2017learning} in 2017. This was the first attempt to automatically construct and tune DA policies according to provided data. The parameterization in TANDA inspired the design of search space in AutoAugment (AA) \cite{cubuk2019autoaugment}, and provided a standard problem formulation to the AutoDA field. AA used Reinforcement Learning (RL) to perform the augmentation search. During the search, augmentation policies were sampled via a Recurrent Neural Network (RNN) controller and then used for model training. Instead of directly searching on the target data, AA created a subset out of original training set as a proxy task. The evaluation of augmentation policies were also conducted on a simplified network instead of the final classification model. Unfortunately, searching in AA requires thousands of GPU hours to complete even under reduced setting. 

With the establishment of augmentation search space, efficiency problems have become the focus of later AutoDA works. Fast AutoAugment (Fast AA) \cite{lim2019fast} is one of the most popular improved versions of the original AA. Instead of RNN, Fast AA applies Bayesian optimization to sample the next augmentation policy to be evaluated, which greatly reduces the search cost. Additionally, Fast AA firstly uses density matching for policy evaluation, which completely eliminates the need for repeated training. Another approach to improve search efficiency is via parallel computation. Population Based Augmentation (PBA) \cite{ho2019population} adopts Population Based Training to optimize the augmentation policy using several subsets of the target data simultaneously. The search goal in PBA is also slightly different than previous approaches. PBA aims to find a dynamic schedule during model training, rather than a static policy. Both Fast AA and PBA substantially reduce the complexity of the AA algorithm, and maintain comparable performance at the same time. However, there is still an expensive searching phase in these models especially when faced with large datasets or complicated models, which inevitably leads to poor efficiency. 

To further enhance the scalability of AutoDA models, techniques such as gradient-based hyper-parameter optimization have been explored recently. AutoDA based on gradient is usually achieved by various gradient approximators to estimate the gradient of augmentation hyper-parameters with regard to model performance. This process ensures the hyper-parameters can be differentiated and hence optimized along with the model training. Adversarial AutoAugment (AAA) \cite{zhang2019adversarial} and Online Hyper-parameter Learning AutoAugment (OHL-AA) \cite{lin2019online} apply the REINFORCE gradient estimator \cite{williams1992simple} to achieve gradient approximation. Other gradient estimators are also applicable in AutoDA. For example, DARTS \cite{liu2018darts} is employed in Faster AutoAugment (Faster) \cite{hataya2020faster} and Differentiable Automatic Data Augmentation (DADA) \cite{li2020dada}. Using the same policy model as in Fast AA \cite{lim2019fast}, OHL-AA optimizes augmentation policies in an online fashion during model training. There is no separate stage for searching in OHL-AA. Instead, it adopts a bi-level optimization framework, where the algorithm updates the weights of the classification model and hyper-parameters of augmentation policy at the same time. This scheme significantly reduces the search time. Similarly, there are two optimization objectives in AAA, one of which is the minimization of training loss, and the other is the minimization of adversarial loss \cite{zhang2019adversarial}. Two objectives are optimized simultaneously in AAA, providing a much more computationally-affordable solution. 

Even though gradient-based approaches are more efficient in comparison to vanilla AA, these methods are still based on an expensive policy search. The bi-level setting also increases the complexity of the model training stage. Recent advancements in AutoDA aim to further enhance the efficiency of augmentation design by excluding the need for search. Proposed in 2020, RandAugment (RA) \cite{cubuk2020randaugment} reparameterizes the classical search space. It replaces the individual parameter for each transformation with two global variables. A simple grid search is performed in RA to optimize two hyper-parameters. The findings in RA not only suggest that the traditional search phase may not be necessary, but also indicate that the search using surrogate models could be sub-optimal. It was found that the effectiveness of the DA policy was relevant to the size of the model and training set, thus challenges all previous approaches based on proxy tasks. Another AutoDA model that does not rely on searching is UniformAugment (UA) \cite{lingchen2020uniformaugment}. UA further simplifies the augmentation space through invariance theory. The authors hypothesize an approximate invariant augmentation space. Any augmentation policy sampled from that space could lead to similar model performances, thus completely removing the search phase. Despite the promising speed, the model performance is a bottleneck in these search-free methods. Neither approach is able to make significant progress on model accuracy when compared with previous approaches. To apply AutoDA techniques in practice, further research and experiments are required. 


\section{Automated Data Augmentation Techniques}
\label{Sec: autoda}

This section aims at introducing the basic concepts and terminologies of Automated Data Augmentation (AutoDA) techniques. In general, finding an optimal DA policy is formulated as a standard search problem in most works \cite{cubuk2019autoaugment, ho2019population, lim2019fast, hataya2020faster, lin2019online, li2020dada}. A standard AutoDA model consists of three major components: a search space, a search algorithm and an evaluation function. In this section, the functionalities and relationships of three component are discussed. We also describe how to assess the proposed AutoDA models, including two different evaluations based on direct and indirect approaches. Lastly, we introduce several commonly used datasets and benchmarks for comparative analysis.

\subsection{Key Components}

For image classification tasks, the major objective of AutoDA models is to achieve the best classification accuracy using an optimal DA policy automatically learned from a given dataset. Inspired by the DA strategy modelling in Transformation Adversarial Networks for Data Augmentations (TANDA) \cite{ratner2017learning}, AutoAugment (AA) \cite{cubuk2019autoaugment} is considered to be the first work that attempted to automate the augmentation policy search. AA formulates the AutoDA task as a search problem, and provided basic parameterization for the search space. The parameterization in AA is largely adopted in later AutoDA works, and is regarded as the de facto standard \cite{lim2019fast, ho2019population, wei2020circumventing}. Specifically, there are three key components within a standard AutoDA formulation:

\begin{definition}[\textbf{Search Space}]
is regarded as the domain of DA policies to be optimized where all candidate solutions via augmentation hyper-parameters are defined. 
\end{definition}

\begin{definition}[\textbf{Search Algorithm}]
is used to retrieve augmentation policies from the search space and to sample the next search point based on a reward signal returned by an evaluation function.
\end{definition}

\begin{definition}[\textbf{Evaluation Function}]
is the procedure of assessing or ranking sampled DA policies by assigning reward values. This usually relies on the training of the classification model.
\end{definition}

\subsubsection{Search Space}

The search space defines how DA policies are formed for subsequent searches. Specifically for image classification tasks, the augmentation policy refers to a composition of several image operations, which can be described by augmentation hyper-parameters. Generally, a complete augmentation policy consists of multiple sub-policies, each of which is used to augment one training batch. A sub-policy is composed of several basic image Transformation Functions (TFs). An augmentation policy is usually parameterized by two hyper-parameters: the application probability and the operation magnitude. The probability describes the probability of applying a certain transformation function on input images, while the magnitude determines the strength of the operation. Each TF within a DA policy is associated with a pair of probability and magnitude hyper-parameters. Depending on the choice of search or optimization algorithm, the formulation of the search space can vary greatly. For example, some works completely re-parameterize the search space to reduce the search complexity \cite{cubuk2020randaugment, lingchen2020uniformaugment}. However, the DA policy parameterization proposed in AA \cite{cubuk2019autoaugment} has been widely used in later works with little or no modification.

\subsubsection{Evaluation Functions}

The evaluation of augmentation policies is conducted from two perspectives, including the effectiveness and safety. The former feature emphasizes the impacts of DA on final classification results, while the safety feature focuses on the label preservation of the transformed data. Generally, the efficacy of augmentations is judged by the performance of classification models based on training loss or accuracy values. Such procedures can also be called direct evaluation functions, since the strength of the DA strategy is directly reflected in how much performance gains this augmentation policy can produce. The higher the classification accuracy, the better the associated DA policy.

Another alternative evaluation is an indirect method, emphasizing the safety feature of data augmentation. Examining the safety of DA policy often resorts to the use of density matching \cite{lim2019fast}. The main objective of density matching is to match the distribution of the augmented data to the original training data. The basic idea behind this indirect evaluation is to treat the transformed images as missing data points of the input data, thereby improving the generalizability of the classification model. Smaller density differences indicate higher similarity of data distributions, which can lead to better augmentation strategies. Using density matching, the policy evaluation does not require back-propagation of the model training. Such algorithms can be regarded as the indirect evaluation function of AutoDA tasks. 

\subsection{Overall Workflow}

\begin{figure}[h]
  \centering
  \includegraphics[width=0.7\linewidth]{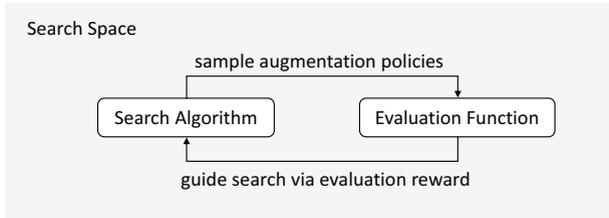}
  \caption{General workflow of a standard AutoDA model involving three key components.}
  \label{Fig: components}
\end{figure}

The relationship between these components is depicted in Fig. \ref{Fig: components}. Firstly, the AutoDA model specifies the parameterization of DA policies for the given task, providing a finite number of potential solutions to be searched and evaluated. Within the defined search space, the search algorithm then samples DA policies and passes the candidates to the evaluation section. In earlier AutoDA works, augmentation policies were sampled one by one \cite{cubuk2019autoaugment, naghizadeh2020greedy}, while later approaches tend to employ multi-threaded processes, sampling multiple candidates and evaluating them in a distributed fashion. This substantially improves the search efficiency. After a DA policy is selected by the search algorithm, it is then rated by the evaluation function to compute the reward signal. Each augmentation strategy is associated with a reward value, indicating its effectiveness in improving model performance. Finally, the reward information is used to update the search algorithm, guiding the sampling of the next DA policy to be evaluated. The entire search recursion process ends when the optimal policy is found. This can be determined by examining the difference in performance gain between the current search point and the previous point. However, the stopping criteria might lead to excessive searching with little improvement, especially in the later phases, resulting in a waste of resources. In most practical implementations of AutoDA algorithms, the search algorithm will stop when a self-defined stopping condition is fulfilled, for example after a certain number of search epochs \cite{niu2019automatically, lim2019fast, ho2019population}.

\subsection{Two Stages of AutoDA}

The standard AutoDA pipeline can be divided into two major stages: 

\begin{definition}[\textbf{Generation stage}]
is the process of generating the optimal augmentation policy when given certain datasets. A DA policy is typically described by a sequence of augmentation hyper-parameters. Usually, the final DA solution is generated by a search or optimization algorithm, which samples various candidate strategies from the defined search space, and relies on a evaluation function to assess efficacy of the searched policies. 
\end{definition}

\begin{definition}[\textbf{Application stage}]
is the process of applying the policy learned in the generation stage. This is done by augmenting the target dataset using the obtained DA policy to artificially increase both the data quantity and variety, and then train the classification model on the transformed training set. 
\end{definition}

With the aim of finding the best augmentation strategy for the target dataset, a typical AutoDA problem is mainly solved in the policy generation stage. The best DA policy here specifically refers to the hyper-parameter setting that can maximize the classification model accuracy or minimize the training loss in the later phase, i.e. it can best solve the classification task in application phase. We identify several criterion used in published studies to determine the completion of policy generation: 

\begin{enumerate}[1.]
    \item the sampled DA policy can help train the classification model to achieve the highest accuracy scores; 
    \item the sampled DA policy can provide comparable performance gains to the optimal policy; or
    \item a certain number of training/searching epochs has been completed. 
\end{enumerate}

Theoretically, the policy generation stage can only end when the first criterion is achieved, i.e. the sampled policy is evaluated to be the optimal augmentation strategy for the given dataset. However, it is often impractical to thoroughly explore the entire search space to identify the best DA policy. A potential solution is to set a specific threshold for model accuracy or training loss to help decide whether the policy is optimal. However, in application scenarios, the optimal strength of data augmentation for classification models is often unknown. It is therefore tricky to set such thresholds as success criteria. 

An alternative strategy to stop the generation phase is to relax the optimal criteria. In other words, if the sampled policy can produce comparable improvement in model performance to the optimal DA strategy, it is considered to be optimal. This idea has been widely adopted in many existing AutoDA works \cite{lim2019fast, gudovskiy2021autodo, naghizadeh2020greedy, naghizadeh2021greedy}. It can be implemented by using the performance difference. For example, if the difference in performance gains between the sampled policy and the best rewarded one is smaller than a certain value, then this policy can be treated as the final output of the generation stage \cite{naghizadeh2020greedy, naghizadeh2021greedy}. A more popular alternative is to use density matching. Instead of directly training the classification model, density matching compares the distribution/density of the original data and the augmented samples. The assumption of density matching is that the optimal DA policy can best generalize the classification model by matching the density of given data with the density of the transformed data \cite{lim2019fast, gudovskiy2021autodo}. 

In practice, the most commonly used stopping criteria is to manually decide the search limit. Once the training has been conducted after a certain number of epochs, policy generation will be forced to stop and output the DA policy with the best model performance so far. The selection of epoch number usually depends on the available computational resources as well as the complexity of the given task. There is no universal agreement on the stopping criteria.

\subsection{Datasets}

This section aims at providing a brief overview of the datasets employed in the considered approaches. Annotated datasets are generally used as benchmarks to provide a fair comparison among different AutoDA algorithms and architectures. Furthermore, the growth in size of data and complexity of application scenarios increases the challenge, resulting in constant development of new and improved techniques. 

The most used datasets for the task of automated augmentation search are: (i) CIFAR-10/100 \cite{krizhevsky2009learning}, (ii) SVHN \cite{netzer2011reading}, (iii) ImageNet \cite{russakovsky2015imagenet}. CIFAR stands for Canadian Institute for Advanced Research. CIFAR-10 and CIFAR-100 share the same name as both are used for CIFAR research, while the numbers specifies the total number of classes in the dataset. SVHN refers to Street View House Numbers (SVHN). ImageNet is used in the ImageNet Large Scale Visual Recognition Challenge (ILSVRC) \cite{russakovsky2015imagenet}. The characteristics of each dataset are shown in Table \ref{tab: datasets}, while their statistics are summarized in Table \ref{tab: data_stats}. 

\begin{table}[h]
\begin{center}
\begin{minipage}{240pt}
\caption{Main characteristics of datasets for AutoDA tasks}\label{tab: datasets}%
\begin{tabular}{@{}lllll@{}}
\toprule
Dataset & Classes  & Images per class & Image size & Year\\
\midrule
CIFAR-10         & 10      & 6,000    & $32 \times 32$ & 2009 \\ 
CIFAR-100        & 100     & 600     & $32 \times 32$ & 2009 \\ 
SVHN             & 10      & 630,420  & $32 \times 32$ & 2011 \\ 
ImageNet & varying\footnotemark[1] &varying\footnotemark[1] & varying\footnotemark[1] & 2009 \\ 
\botrule
\end{tabular}
\footnotetext[1]{The complete ImageNet is too large for an augmentation search. Instead, a trimmed subset of ImageNet is usually used in AutoDA works, while such set is constructed differently in each work.}
\end{minipage}
\end{center}
\end{table}

\begin{table}[h]
\begin{center}
\begin{minipage}{\textwidth}
\caption{Statistics of datasets for AutoDA tasks}\label{tab: data_stats}
\begin{tabular*}{\textwidth}{@{\extracolsep{\fill}}lcccccc@{\extracolsep{\fill}}}
\toprule%
& \multicolumn{3}{@{}c@{}}{Number of images} & \multicolumn{3}{@{}c@{}}{ Number of images per class} \\\cmidrule{2-4}\cmidrule{5-7}%
Dataset & Train & Test & Extra \footnotemark[1] & Train & Test & Extra \footnotemark[1] \\
\midrule
CIFAR-10  & 50,000  & 10,000 & - & 5,000 & 1,000 & - \\
CIFAR-100  & 50,000 & 10,000 &- & 500 & 100 & - \\
SVHN  & 73,257 & 26,032  & 531,131 & 5,000 - 14,000 & 1,800 - 5,000 & 35,000 - 90,000 \\
\botrule
\end{tabular*}
\footnotetext[1]{Extra set consists of the same type of annotated images as training data, but with slightly lower quality. These data are often used as extra training samples for the model. Extra set is only available in SVHN data \cite{netzer2011reading}.}
\end{minipage}
\end{center}
\end{table}

The CIFAR and ImageNet dataset are published at the same year, both present standard image classification task and are commonly used in computer vision researches. However, ImageNet is much larger than the CIFAR series in scale and diversity. There are over $5,000$ different categories in the original ImageNet set, with $3.2$ million images that have been hand-annotated \cite{deng2009imagenet}. For the AutoDA search problem, the enormous quantity of ImageNet data might require significant amounts of computational resources. Training on the complete ImageNet is usually infeasible in practice. Instead, it is often more suitable to use a reduced ImageNet subset for the target task. Additionally, the distribution of instances among different classes can also vary considerably, which can decrease the performance of AutoDA models. To address these issues, each AutoDA work that conducts experiments using ImageNet data uses a distinctive trimming method to set up a smaller and cleaner subset for model evaluation. Nevertheless, due to the diversity of data and imbalanced class distribution, the classification on ImageNet subset is still considered to be a relatively difficult task when compared with other datasets (for augmentation search). In AutoDA works, the reduced ImageNet datasets are constructed differently, with varying sizes and class numbers. Each of them will be described in the works where they are employed.

The CIFAR series consists of much fewer categories, designated by their number \cite{krizhevsky2009learning}. The CIFAR-10 dataset consists of $60,000$ $32\times32$ colour images in total. The data distribution among classes in CIFAR-10 is more controlled and unified. $60,000$ images are evenly distributed into $10$ classes, providing $6,000$ images per class. The splitting ration of train:test data is $5:1$. In CIFAR-10 dataset, the test batch contains $10,000$ images, randomly selected from the full dataset, but each class contains exactly $1,000$ images. The training set contains the remaining $50,000$ instances. The formulation of CIFAR-100 dataset is similar to CIFAR-10, except there are $100$ classes in CIFAR-100, each of which comprises $600$ images. The train:test ratio is also $5:1$, providing $500$ training images and $100$ test images per class. With a more balanced data distribution and limited class number, CIFAR data is usually more suitable to benchmark proposed AutoDA algorithms. 

SVHN refers to Street View House Numbers. It is also collected from real-world scenarios, widely used for deep learning related researches. Similar to MNIST data \cite{lecun1998gradient}, images in SVHN are also digits, cropped from house numbers in Google Street View images \cite{netzer2011reading}. The major task of SVHN is to recognize numbers in natural scene images. There are $10$ categories in total, each of which represents one digit, e.g. digit 1 has label 1. In SVHN, there are $73,257$ digits for training, $26,032$ digits for testing, and $531,131$ additional data tiems that can be used as extra training data. In contrast to previous datasets, SVHN specifically emphasizes the pictures of digits and numbers. This might reveal the relationship between DA selection strategy and data types. However, unlike CIFAR, the SVHN data distribution among classes is biased. There are more 0 and 1 digits present in the data, resulting in a skewed class distribution in both training and test set. Seen in Table. \ref{tab: data_stats}, for SVHN, the number of images per class ranges from $5,000$ to $14,000$. Such imbalance can be considered as a challenge to better assess AutoDA models from different perspectives.


\section{Taxonomy of Image AutoDA Methods}
\label{Sec: taxonomy}

Table \ref{tab: summary} shows a summary of primary works in the AutoDA field. The column \textit{Key technique} describes the most important technique adapted in each AutoDA model to formulate augmentation search problems. These methods are usually borrowed from other ML-related field, such as NAS or hyper-parameter optimization. \textit{Policy optimizer} indicates the algorithm or controller that is used to optimize or update the augmentation policy during the search. These AutoDA approaches are classified into two major types based on the stage involved to solve the classification task using the learned DA policy, namely one-stage or two-stage. Additionally, from the perspective of hyper-parameter optimization, these methods can be further categorized into three classes: gradient-based, gradient-free and search-free. Table \ref{tab: summary} provides a categorization that projects the underlying optimization algorithm used by each of the methods.

Based on the application sequence of the two stages of the AutoDA model, we classify all existing works into two major categories: one-stage and two-stage approaches. Two-stage approaches conduct both the generation and application respectively. In a typical two-stage method, the optimal augmentation policy is generated according to the task dataset from the first stage. After that, the learned augmentation strategies is applied on the training set to train the discriminative model. There are two separate stages required when utilizing a two-stage algorithm. The one-stage approaches combine the generation and application together through the use of gradient approximation methods. By estimating the gradient of a DA policy with regard to model performance, one-stage approaches are able to optimize augmentation policy and classification model simultaneously. As a result, they can obtain the final results and trained model through a single run. Such algorithms usually achieve better efficiency than traditional two-stage approaches. 

\begin{sidewaystable}
\sidewaystablefn%
\begin{center}
\begin{minipage}{\textheight}
\caption{Summary table of major AutoDA works reviewed for image classification task}\label{tab: summary}
\begin{tabular*}{\textheight}{@{\extracolsep{\fill}}lccccccc@{\extracolsep{\fill}}}
\toprule%
& & & \multicolumn{2}{@{}c@{}}{Stage \footnotemark[2]}& \multicolumn{3}{@{}c@{}}{Policy optimization \footnotemark[3]} \\\cmidrule{4-5}\cmidrule{6-8}%
Method	& Key technique & Policy optimizer \footnotemark[1] & two-stage & one-stage & gradient-free & gradient-based & search-free \\
\midrule
\multirow{2}{*}{TANDA\footnotemark[4]\cite{ratner2017learning}} & Generative & Long short-term & $\checkmark$ & & $\checkmark$ & & \\ &adversarial network & memory network \\
\midrule
\multirow{2}{*}{AA\footnotemark[5]\cite{cubuk2019autoaugment}} & Reinforcement & Recurrent neural & $\checkmark$ & & $\checkmark$ & & \\ &learning (RL) & network controller \\
\midrule
\multirow{2}{*}{AWS\footnotemark[6]\cite{tian2020improving}} & Weight sharing; & Proximal policy & $\checkmark$ & & $\checkmark$ & & \\ & RL & optimization \\

\botrule
\end{tabular*}
\footnotetext[1]{Policy optimizer is the algorithm or controller used by AutoDA model to update augmentation strategies during the policy generation stage.}
\footnotetext[2]{Classification based on the application sequence of two stages involved. Policy generation and application are performed simultaneously in an one-stage approach, but sequentially in a two-stage method.}
\footnotetext[3]{Classification based on the type of policy optimization. Gradient-free methods optimize augmentation policies without approximating the gradients of policy hyper-parameters. Gradient-based methods estimates such gradients for policy optimization. Search-free methods re-parameterize the search space to exclude the need of search.}
\footnotetext[4]{Transformation Adversarial Networks for Data Augmentations}
\footnotetext[5]{AutoAugment}
\footnotetext[6]{Augmentation-WiseWeight Sharing}
\end{minipage}
\end{center}
\end{sidewaystable}

\begin{sidewaystable}
\sidewaystablefn
\renewcommand\thetable{3}
\begin{center}
\begin{minipage}{\textheight}
\caption{(continued)}\label{tab: summary2}
\begin{tabular*}{\textheight}{@{\extracolsep{\fill}}lccccccc@{\extracolsep{\fill}}}
\toprule%
& & & \multicolumn{2}{@{}c@{}}{Stage \footnotemark[2]}& \multicolumn{3}{@{}c@{}}{Policy optimization \footnotemark[3]} \\\cmidrule{4-5}\cmidrule{6-8}%
Method	& Key technique & Policy optimizer \footnotemark[1] & two-stage & one-stage & gradient-free & gradient-based & search-free \\
\midrule
\multirow{2}{*}{GAA\footnotemark[7]\cite{naghizadeh2020greedy, naghizadeh2021greedy}} & Greedy breadth & Breadth-first & $\checkmark$ & & $\checkmark$ & & \\ &first search & search \\
\midrule
\multirow{2}{*}{PBA\footnotemark[8]\cite{ho2019population}} & Population-based & Truncation & $\checkmark$ & & $\checkmark$ & & \\ &training & selection \\
\midrule
\multirow{2}{*}{Fast AA\footnotemark[9]\cite{lim2019fast}} & Density matching; & Bayesian & $\checkmark$ & & $\checkmark$ & & \\ & RL & optimization \\
\midrule
\multirow{2}{*}{PAA\footnotemark[10]\cite{lin2021local}} & Multi-agent & Advantage actor& $\checkmark$ & & $\checkmark$ & & \\ & RL & critic algorithm\\
\midrule
\multirow{2}{*}{AA-KD\footnotemark[11]\cite{wei2020circumventing}} & Knowledge & - & $\checkmark$ & & $\checkmark$ & & \\ & distillation & \\

\botrule
\end{tabular*}
\footnotetext[7]{Greedy AutoAugment}
\footnotetext[8]{Population-Based Augmentation}
\footnotetext[9]{Fast AutoAugment}
\footnotetext[10]{Patch AutoAugment}
\footnotetext[11]{AutoAugment with Knowledge Distillation}
\end{minipage}
\end{center}
\end{sidewaystable}

\begin{sidewaystable}
\sidewaystablefn
\renewcommand\thetable{3}
\begin{center}
\begin{minipage}{\textheight}
\caption{(continued)}\label{tab: summary3}
\begin{tabular*}{\textheight}{@{\extracolsep{\fill}}lccccccc@{\extracolsep{\fill}}}
\toprule%
& & & \multicolumn{2}{@{}c@{}}{Stage \footnotemark[2]}& \multicolumn{3}{@{}c@{}}{Policy optimization \footnotemark[3]} \\\cmidrule{4-5}\cmidrule{6-8}%
Method	& Key technique & Policy optimizer \footnotemark[1] & two-stage & one-stage & gradient-free & gradient-based & search-free \\
\midrule
\multirow{2}{*}{Faster AA\footnotemark[12]\cite{hataya2020faster}} & Gradient estimation; & Stochastic & $\checkmark$ & & & $\checkmark$ & \\ & Density matching & gradient descent \\
\midrule
\multirow{2}{*}{RA\footnotemark[13]\cite{cubuk2020randaugment}} & Search space & Grid search & $\checkmark$ & & & & $\checkmark$ \\ & reparameterization & \\
\midrule
\multirow{2}{*}{UA\footnotemark[14]\cite{lingchen2020uniformaugment}} & Augmentation & Uniform & $\checkmark$ & & & & $\checkmark$ \\ & invariance & sampling \\
\midrule
\multirow{2}{*}{OHL-AA\footnotemark[15]\cite{lin2019online}} & Gradient estimation & Stochastic & & $\checkmark$ & & $\checkmark$ & \\ &  & gradient descent \\
\midrule
\multirow{2}{*}{Adv AA\footnotemark[16]\cite{zhang2019adversarial}} & Gradient estimation; & Recurrent & & $\checkmark$ & & $\checkmark$ & \\ & Adversarial learning & neural network \\

\botrule
\end{tabular*}
\footnotetext[12]{Faster AutoAugment}
\footnotetext[13]{RandAugment}
\footnotetext[14]{UniformAugment}
\footnotetext[15]{Online Hyper-parameter Learning for Auto-Augmentation}
\footnotetext[16]{Adversarial AutoAugment}
\end{minipage}
\end{center}
\end{sidewaystable}

\begin{sidewaystable}
\sidewaystablefn
\renewcommand\thetable{3}
\begin{center}
\begin{minipage}{\textheight}
\caption{(continued)}\label{tab: summary3}
\begin{tabular*}{\textheight}{@{\extracolsep{\fill}}lccccccc@{\extracolsep{\fill}}}
\toprule%
& & & \multicolumn{2}{@{}c@{}}{Stage \footnotemark[2]}& \multicolumn{3}{@{}c@{}}{Policy optimization \footnotemark[3]} \\\cmidrule{4-5}\cmidrule{6-8}%
Method	& Key technique & Policy optimizer \footnotemark[1] & two-stage & one-stage & gradient-free & gradient-based & search-free \\
\midrule
\multirow{2}{*}{DADA\footnotemark[17]\cite{li2020dada}} & Unbiased & Stochastic & & $\checkmark$ & & $\checkmark$ & \\ & gradient estimator & gradient descent \\
\midrule
\multirow{3}{*}{AutoDO\footnotemark[18]\cite{gudovskiy2021autodo}} & Gradient estimation; & Stochastic & & & & & \\ & Loss reweighting; & gradient descent & & $\checkmark$ & & $\checkmark$ & \\ & Soft labelling & & & & & &\\

\botrule
\end{tabular*}
\footnotetext[17]{Differentiable Automatic Data Augmentation}
\footnotetext[18]{Automated Dataset Optimization}
\end{minipage}
\end{center}
\end{sidewaystable}

\subsection{Two-stage Methods}

\begin{figure}[h]
  \centering
  \includegraphics[width=\linewidth]{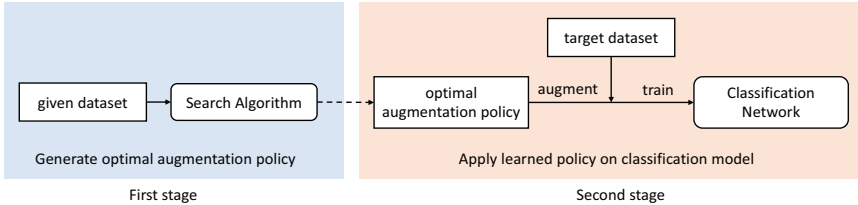}
  \caption{Overall framework of two-stage AutoDA approaches.}
  \label{Fig: two_stage}
\end{figure}

There are two steps involved in applying AutoDA to discriminative tasks in the imaging domain. Generally, an AutoDA model searches for the optimal augmentation strategy and then applies the obtained policy on the target data for model training. Due to the separate processes of searching and training, this kind of approach is described as two-stage in this paper. The general framework of a typical (two-stage) approach is displayed in Fig. \ref{Fig: two_stage}. In the first stage, given a specific dataset, the search algorithm looks for the best composition of image transformation functions, also known as the DA policy. The generation stage ends once the optimal policy is identified by the evaluation function or the searching reaches a given time limit. In the second stage, the learnt policy is applied on the target training set  - ideally with additional data of increased quantity and targeted variety. Then the augmented training samples are fed into the classification model for final training. 

The algorithm used to find the best scheme for data augmentation has been explored in a wide range of existing works. We categorize them into three different classes according to the problem formulation. Some works treat augmentation searching as a standard gradient-free optimization problem \cite{ratner2017learning, cubuk2019autoaugment, ho2019population, lim2019fast, tian2020improving, naghizadeh2020greedy, wei2020circumventing}, whilst other methods approach it from the gradient perspective by means of various gradient approximation algorithms \cite{hataya2020faster}. Other options re-parameterize the entire AutoDA problem in a way that eliminates the need for searching - so called search-free approaches. 

\subsubsection{Gradient-free}

Gradient-free approaches search for the best parameters of the augmentation policy based on model hyper-parameter optimization without gradient approximation. Intuitively, such optimizations can be accomplished by selecting several values for each hyper-parameter, completing a model training for each combination on the target task, and then computing the evaluation metrics of the model performance using all hyper-parameter values. The first attempt to automate such a search process was Transformation Adversarial Networks for Data Augmentations (TANDA) \cite{ratner2017learning}, which utilizes a Generative Adversarial Network (GAN) architecture at its core. The objective of the generator is to propose appropriate sequences of arbitrary augmentation operations, which are then sent to a discriminator for effectiveness assessment. The problem formulation in \cite{hu2019learning} motivated the deep learning community to explore other methods such as AA \cite{cubuk2019autoaugment}. AA inherits the augmentation sequence modeling in \cite{lemley2017smart}, but applies a different strategy based on Reinforcement Learning (RL). Several possible augmentation policies are sampled via a Recurrent Neural Network (RNN) controller, that are then assessed through training a simplified child model instead of given classification model. Despite its promising performance in terms of model improvement, AA has a non-negligible limitation, which is the extremely low efficiency. It can take up to $15,000$ GPU hours to complete a search over ImageNet data. Even with the smallest CIFAR-10 set, AA still requires thousands of GPU hours to complete a single run.  

The majority of the later works on this topic aim to contribute to efficiency enhancements and computational cost reductions. For example, \cite{tian2020improving} utilizes a similar reinforcement learning method, but slightly modifies the search procedure by sharing the same augmentation parameters from earlier stages. Such auto-augment techniques can be further improved through the application of advanced evolutionary algorithms such as Population Based Training (PBT) \cite{jaderberg2017population}. Some simple searching algorithms have also been found to be beneficial to accelerate the first stage. For instance, \cite{naghizadeh2020greedy, naghizadeh2021greedy} replace the original RNN controller with a traditional Greedy Breadth First Search algorithm to simplify the process, and therefore reduce the overall computation cost. In addition to the selection of the search algorithm, modification of the evaluation function can also greatly reduce the computational demands. A landmark work in this direction is Fast AutoAugment (Fast AA) \cite{lim2019fast}, which takes advantage of the variable kernel density \cite{terrell1992variable} and proposes an efficient density matching algorithm as a substitute. In the AutoDA context, the data density represents the overall distribution of data. Instead of a training classification model, density matching evaluates DA policies by comparing the distribution of the original training set and the transformed data. Such algorithms eliminate the need for re-training the model and hence result in a significant efficiency boost. 

Another approach is to focus on the effectiveness or precision of the learned augmentation policy. All of the aforementioned methods focus on the resources and time consumption of the search phase. Not much progress has been made in terms of the improvement of classification accuracy. To fill this gap, \cite{lin2021local} proposes a more fine-grained Patch AutoAugment (PAA) technique which optimizes the augmentation transformations targeted to local regions of images rather than the whole image. Other state-of-the-art methods in the Network Architecture Search (NAS) field help to increase the augmentation precision. One example is Knowledge Distillation (KD) \cite{hinton2015distilling} as used in \cite{wei2020circumventing}. 

\subsubsection{Gradient-based}

In contrast to gradient-free algorithms, approaches that approximate the gradient of hyper-parameters to be searched are referred to as gradient-based optimizations. So far, the only two-stage approach based on gradients is Faster AutoAugment (Faster AA) \cite{hataya2020faster}. This achieves a more efficient augmentation search for image classification tasks than prior methods including AutoAugment \cite{cubuk2019autoaugment}, Fast AA \cite{lim2019fast} and PBA \cite{ho2019population}. The authors of Faster AA adapt an innovative gradient approximation method, namely Relaxed Bernoulli distribution \cite{jang2016categorical}, to relax the non-differentiable distributions of hyper-parameters and use their gradients as input to a standard optimization algorithm. The consecutive two phases can therefore be done within a single pass. Faster AA model jointly optimizes the hyper-parameters of the augmentation policy (i.e. generation phase) and weights of the classification model (i.e. application phase). The simplification of the policy search space significantly reduces the search cost especially when compared to previous algorithms whilst maintaining the performance. What should be emphasized here is that the model trained during the search in Faster AA is actually abandoned later. To get the final classification result, the learned policy is applied to train the target classification model again. Hence there are still two stages involved in the Faster AA scheme. 

\subsubsection{Search-free}

Despite the advantages of the aforementioned approaches, the added complexity of standard two-stage AutoDA methods might need prohibitive computing resources, for example the original implementation of AA in \cite{cubuk2019autoaugment}. Subsequent works mainly aim to accelerate the search cost \cite{lim2019fast, ho2019population} and utilize gradient approximation \cite{hataya2020faster}. However, these approaches still require an expensive search stage, which usually relies on a simplified proxy task to alleviate efficiency issues. This setting presumes that the learned DA policy based on the proxy task can be transferred to the larger target dataset. However, such assumptions are challenged in \cite{cubuk2020randaugment}. According to the findings in \cite{cubuk2020randaugment}, a proxy task might produce sub-optimal DA policies. 

To solve the aforementioned problems, several works aim to re-formulate the search problem in AutoDA. These approaches are acknowledged as search-free methods due to the complete exclusion of the search phase. By challenging the optimality of traditional AutoDA methods, search-free approaches re-parameterize the entire search space, resulting in ea small number of hyper-parameters, which can be manually adjusted. Therefore, there is no need to conduct the search anymore \cite{gudovskiy2021autodo}. Additionally, it is now feasible to directly learn from the full target dataset instead of a reduced proxy task. Therefore, AutoDA models may learn an augmentation policy more tailored to the task of interest instead of through small proxy tasks.

Existing works such as \cite{lingchen2020uniformaugment} and \cite{cubuk2020randaugment} both belong to the search-free category. Both approaches completely re-parameterize the entire search space so that there is no need to perform resource-intensive searches at all. RandAugment (RA) replaces the enormous search space with a small search space controlled by only two parameters. Both parameters are human-interpretable such that a simple grid search is quite effective. Inspired by RA, UniformAugment (UA) further reduces the complexity of the search space by assuming the approximate invariance of the augmentation space, where uniform sampling is sufficient. Both methods completely avoid a search phase and dramatically increase the efficiency of AutoDA algorithms while maintaining their performance.

\subsection{One-stage Methods}
\label{Sec: one-stage}


The biggest difference between two-stage and one-stage approaches is the joint optimization process in the latter. Previous approaches in the two-stage category mainly rely on an additional surrogate model for policy sampling. They then evaluate the sampled policies via full training on another classification network. The expensive training and evaluation procedure leads to efficiency bottlenecks of AutoDA techniques. To mitigate this issue, one-stage approaches complete the policy generation and application in one single step, eliminating the need for repetitive model training. In standard one-stage schemes, the weights of the classification network and the hyper-parameters of the augmentation policy are optimized simultaneously. This is implemented by a bi-level optimization scheme \cite{colson2007overview}. 

At the inner level, they seek to optimize the weights of the discriminative networks, whilst at the outer level looking for hyper-parameters that describe the optimal augmentation policy, under which they can obtain the best performed model as solution to the inner problem. Due to the dependency of inner and outer level optimization, the learning of these two goals are conducted in an interleaved way. Specifically, a separate augmentation network is adapted to describe the probability distribution of sampled policies. The parameters of such a policy model are regarded as hyper-parameters, which are updated after a given number of epochs of inner training \cite{lin2019online, zhang2019adversarial}. In this bi-level framework, the distribution hyper-parameters and network weights are optimized simultaneously. The minimization of training loss (inner objective) can be easily achieved through classical Stochastic Gradient Descent (SGD), while the vanilla gradient of outer objective is relatively hard to obtain, as the model accuracy is non-differentiable with regard to augmentation hyper-parameters. Therefore, one-stage AutoDA models need to leverage gradient approximation to estimate such gradients for later optimization. In other words, all one-stage approaches in AutoDA are based on gradients.

\subsubsection{Gradient-based}
As its name suggests, gradient-based models optimize the augmentation policy from the perspective of gradients. The reason it has to rely on gradient approximation is because the original model accuracy is non-differentiable with regard to augmentation policy distribution. Only after the relaxation of distribution, can the gradient of validation accuracy or training loss with regard to hyper-parameters be obtained. There are several advantages of gradient-based approaches. Due to the differentiable accuracy, gradient-based method can directly optimize the hyper-parameters according to the estimated gradient. There is no need to invest a significant amount of time in training child models to test sampled policies. This substantially reduces the workload of policy evaluation. The removal of expensive evaluation procedures also enables the AutoDA algorithm to scale up to even larger datasets and deeper models. The first one-stage AutoDA work based on gradients was Online Hyper-parameter Learning AutoAugment (OHL-AA) \cite{lin2019online} in 2019, based on the REINFORCE gradient estimator \cite{williams1992simple}. The augmentation policy model in OHL-AA is similar to previous works \cite{cubuk2019autoaugment, lim2019fast}, while the original search problem is reformulated as a bi-level optimization task. Published in the same year, Adversarial AutoAugment (AAA) \cite{zhang2019adversarial} employs the same gradient approximator in an adversarial framework, which further eases the efficiency issue. As the NAS technique develops in 2020, Differentiable Automatic Data Augmentation (DADA) \cite{li2020dada} and Automated Dataset Optimization (AutoDO) \cite{gudovskiy2021autodo} use the more advanced DARTS estimator \cite{liu2018darts}.

\section{Two-stage Approaches} 
\label{Sec: two}

In this section, we review two-stage strategies in detail, with focus on the pipeline of the algorithms. We start from the fundamental definition of the augmentation parameters and corresponding search space used in each method. After that, the core algorithms are explored along with their overall workflow. Following that, the major contribution of each method is covered based on experimental results provided in the original paper. Then, we provide a systematic analysis and evaluate the pros and cons of the different two-stage category approaches. Finally, we compare all available two-stage algorithms from the perspective of their accuracy and efficiency, and give suggestions on model selection from a practical application perspective.


\subsection{Gradient-free Optimization}

\subsubsection{Transformation Adversarial Networks for Data Augmentations (TANDA)}

TANDA is considered to be the earliest work supporting automatic discovery of optimised data augmentation policies. Even though other works aimed at automating data augmentation, most of them focused on either creating innovative augmentation algorithms \cite{lemley2017smart}, or generating synthetic training data based on a given set of starting images \cite{tran2017bayesian}. TANDA, on the other hand, used only the basic image operations based on a user's specification, and output a sequence of transformation functions as the final augmentation policy. This made it more relevant to many scenarios with diverse data augmentation demands. 

An augmentation policy is represented as a sequence of image processing operations in \cite{ratner2017learning}. Users need to specify a range of augmentation operations for the TANDA model to select from, which are also called as Transformation Functions (TFs). In order to support various types of TFs, TANDA regards them as black-box functions that ignores application details, and only emphasizes the final effect of such transformations. For instance, a $30^{\circ}$ rotation can be achieved with one single TF, or alternatively it can be split into a combination of three $10^{\circ}$ rotation transformations. The policy modelling in TANDA might not be deterministic or differentiable, but it provides an applicable way of tuning the TF hyper-parameters. 

The major objective of TANDA is to learn a model that can generate augmentation policies composed of a fixed number of TFs. Depending on the types of TFs, the DA policy is modelled in two different ways. The first policy model, namely the straightforward mean field model, assumes each TF in an augmentation policy is selected independently. Therefore, the probability of each operation is optimized individually. Mean field modelling largely reduces the number of learnable hyper-parameters during the search. However, this independent representation can be biased, especially when TFs affect each other. In practical scenarios, a certain image processing operation can lead to totally different effects if applied with other TFs. The actual sequence of TF application also matters when some of the TFs are not commutative. To fully represent the interaction among augmentation TFs, TANDA offers another option to model DA policies, the Long Short-Term Memory (LSTM) network. The LSTM model in TANDA outputs probability distributions over all TFs, which emphases the relationship among searched TFs. 

\begin{figure}[h]
  \centering
  \includegraphics[width=0.7\linewidth]{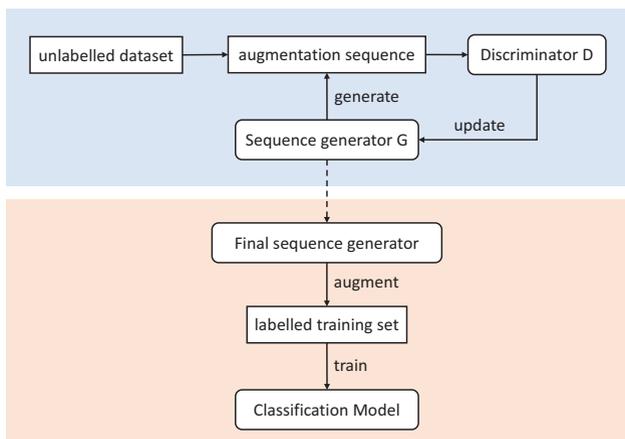}
  \caption{TANDA workflow \cite{ratner2017learning}. Upper/lower sections indicate the policy generation/application stage respectively.}
  \label{Fig: tanda}
\end{figure}

The TANDA model applies standard theGAN architecture, consisting of a generator $G$ and a discriminative model $D$. The general workflow of TANDA is illustrated in Fig. \ref{Fig: tanda}. There are two stages involved in TANDA: policy generation and application. The policy generation phase can be viewed as a classical min-max game in GAN. The goal of sequence generator model $G$ is to sample DA policies that are most likely to fool the discriminator model $D$, while the $D$ tries to distinguish the transformed images out of the original data. This is done by assigning reward values to the input data. Ideally, in-distribution data points will get higher values whereas the images generated via augmentation will be assigned lower rewards. The reward information is then used to update $G$ for the next policy sampling. After the searching is completed, the final generator is used to augment the original training set to better train the classification network. 

There are various advantages of TANDA. Firstly, the performance improvement out of TANDA is convincing. From the experimental results in \cite{ratner2017learning}, TANDA outperforms most contemporaneous heuristic DA approaches. In terms of problem formulation, the LSTM policy model tends to be more effective than mean field representation in most cases, which empirically encourages the sequence modelling in the AutoDA scheme. The proposal of these two policy models is considered to be the most significant contribution of TANDA. The representation of augmentation transformations inspired AutoAugment (AA) \cite{cubuk2019autoaugment}, which also utilized the LSTM model for policy prediction. Furthermore, the positive influence resulted from sequential modelling provides empirical support for later Population-Based Augmentation (PBA) \cite{ho2019population}, which outputs application schedules rather than a fixed policy. The use of unlabelled data is also a favorable characteristic especially for tasks with limited data. Additionally, a trained TANDA model shows a certain degree of robustness against TF mis-specification. In TANDA, there is no limitation on the selection of the TF range or requirement for safety property of available transformations, therefore it is much easier for users to use in practice. More importantly, TANDA is open-source and can be adapted and applied to any task with limited datasets, not only in the imaging domain but also for text data.

\subsubsection{AutoAugment (AA)}

AutoAugment (AA) \cite{cubuk2019autoaugment} is one of the most popular AutoDA approaches. The majority of subsequent works in this field \cite{lim2019fast, ho2019population, hataya2020faster} adapt a similar setup as AA, especially the definition of the search space and policy model. However, the AA algorithm itself does not provide an optimal solution to the policy search problem due to its severe efficiency issues. However, as the authors of AA emphasize, the fundamental contribution of AA lies in the automated approach to DA and the development of the search space, rather than the search strategy. 

AA formulates the automation of DA policy design as a discrete search problem. In AA, an augmentation policy is a composition of $5$ sub-policies, each of which is applied to one training batch. One sub-policy consists of two sequential transformation functions, such as geometric translation, flipping or colour distortion. Each TF in an augmentation policy is described by two hyper-parameters, i.e. the probability of applying this transformation and the magnitude of the application. Inspired by TANDA, the application sequence of these TFs is emphasized. For simplification, the range of probability and magnitude is discrete. The probability is evenly discretized into $11$ values, ranging from $0$ to $1$, whilst the magnitude is selected from positive integers between $1$ to $10$. The $14$ operations implemented in AA are all from standard Python Image Library (PIL). Two additional augmentation techniques, Cutout \cite{devries2017improved} and SamplePairing \cite{inoue2018data}, are also considered due to their effectiveness in classification tasks. Overall, there are $16$ distinct TFs in AA's search space. Finding an augmentation policy via AA thus has $(16\times10\times11)^{10}\approx2.9\times10^{32}$ possibilities. 

\begin{figure}[h]
  \centering
  \includegraphics[width=0.7\linewidth]{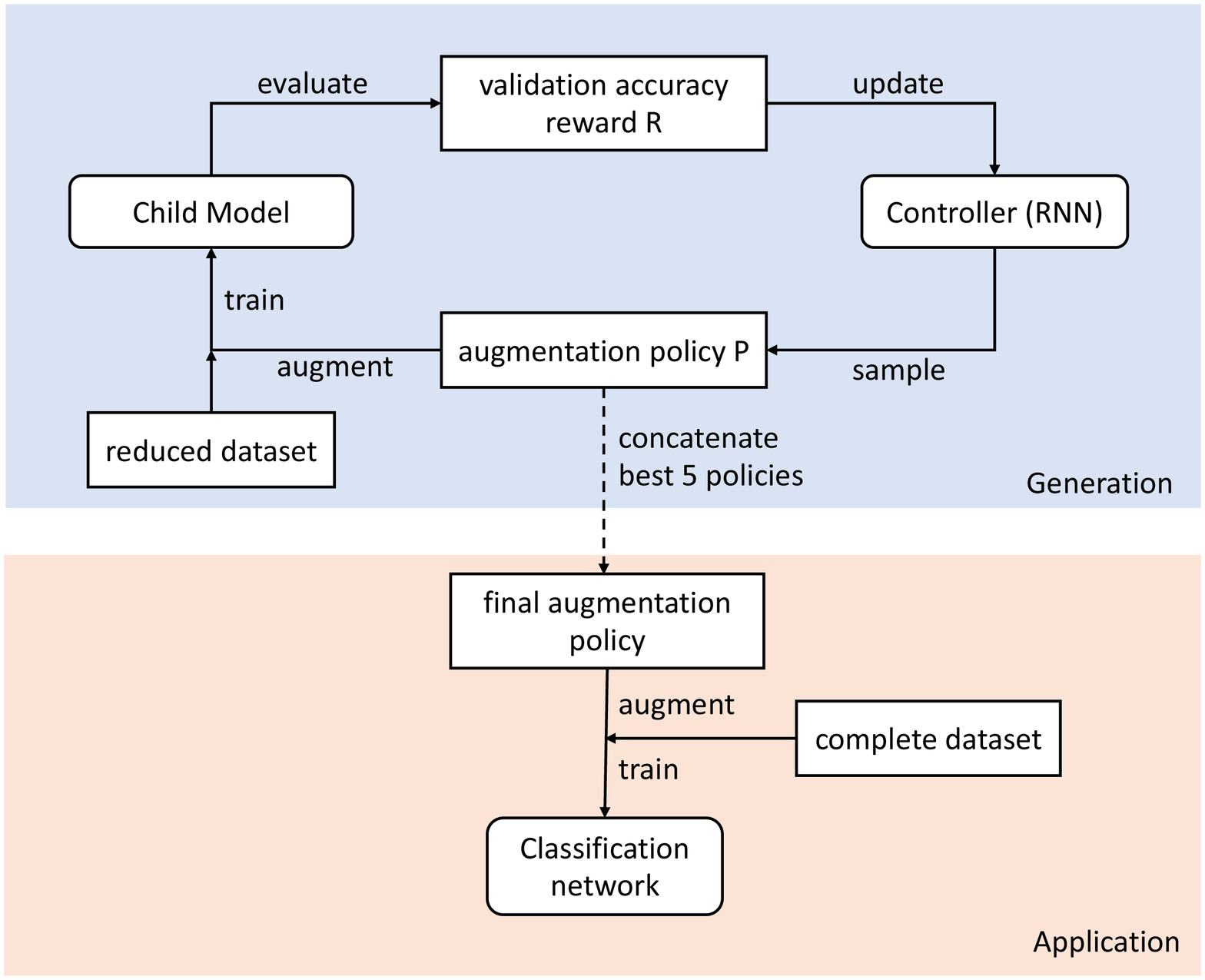}
  \caption{AutoAugment workflow \cite{cubuk2019autoaugment}. Upper/lower sections indicate the policy generation/application stage respectively.}
  \label{Fig: aa}
\end{figure}

To automate the process of constructing DA policy, AA has to search over an enormous search space. It now becomes a discrete search problem using the aforementioned formulation. At a high level, the workflow of AA is displayed in Fig. \ref{Fig: aa}. One of the key components in the AutoDA model is the search algorithm. In the search phase, the search algorithm is used to generate an augmentation policy, which is then evaluated for updates. AA chooses a simple Recurrent Neural Network (RNN) as its search algorithm/controller to sample policy $P$. The evaluation procedure is done through model training, but using reduced data and a simplified model. Such a model is also called a child model, due to its similar but much simpler architecture when compared to the final classification network. After testing the trained child model on a validation set, the validation accuracy is regarded as reward $R$ to update the search controller. Generally, the reward signal $R$ reflects how effective a policy $P$ is in improving the performance of a child model. The training of the child model has to be done multiple times, because $R$ is not differentiable over policy hyper-parameters, i.e. probability and magnitude. 

Through extensive experiments, AA achieves excellent results. It can be directly applied on the target data and achieves competitive model accuracy. Experiments in \cite{cubuk2019autoaugment} report state-of-the-art results for common datasets, including CIFAR-10/100, ImageNet and SVHN. AA not only shows superiority in terms of DA policy design, but also provides the option of transferring the searched policy to other similar data. For example, the augmentation policy leaned on CIFAR-10 can function well on similar data CIFAR-100. There is no need to conduct expensive searches on the later, as the policies discovered by AA are able to be generalized over multiple models and datasets. This is a viable alternative especially when direct search is unaffordable. Another advantage of AA is its simple structure and procedure. The search phase is actually conducted over a subset of data, using a simplified child model. Those simplifications provide direct evaluation of augmentation policies, without the recourse to any complicated approximation algorithms. More importantly, AA standardizes the modelling of the augmentation policy and search space in the AutoDA field. The policy model it designs has been widely acknowledged as the de facto solution. 

However, AA has serious disadvantages. The choice of algorithms in AA can be substantially improved. It applies Reinforcement Learning as the search algorithm, but this selection is made mainly out of convenience. The authors of AA also indicate that other search algorithms, such as genetic programming \cite{real2019regularized} or even random search \cite{bergstra2012random, mania2018simple}, may further improve the final performance. Furthermore, the reduced dataset and simplified model used during the search phase can result in sub-optimal results. According to \cite{cubuk2020randaugment}, the power of an augmentation policy largely depends on the size of model and dataset. Therefore, simplification in AA is likely to introduce bias into the found policy. Additionally, the final policy is formed by a simple concatenation of the $5$ best policies found in the data batch. The application schedule of these policies is not considered in AA. The greatest shortcoming of AA lies in its efficiency. Evaluation of augmentation policies relies on expensive model training. Due to the stochasiticity of DA policies introduced by the probability hyper-parameter, such training has to be conducted for a certain number of epochs till the policy starts to take effect. In most cases, running AA is extremely resource-intensive, which raises timing and cost issues. This also becomes the major challenge for AutoDA tasks and promotes multiple later methods aiming at efficiency improvement.

\subsubsection{Augmentation-wise Weight Sharing (AWS)}

A major reason for the inefficiency of AA is the repeated training process during policy evaluation. To enhance the efficiency of evaluation, some methods \cite{ho2019population, lin2019online, zhang2019adversarial} sacrifice reliability to some extent. On the contrary, Augmentation-wise Weight Sharing (AWS) designs a proxy task based on the weight sharing concept in NAS, proposing a faster but still accurate evaluation process. The augmentation policies found by AWS also achieve competitive accuracy when compared to other AutoDA methods.

\begin{figure}[h]
  \centering
  \includegraphics[width=0.9\linewidth]{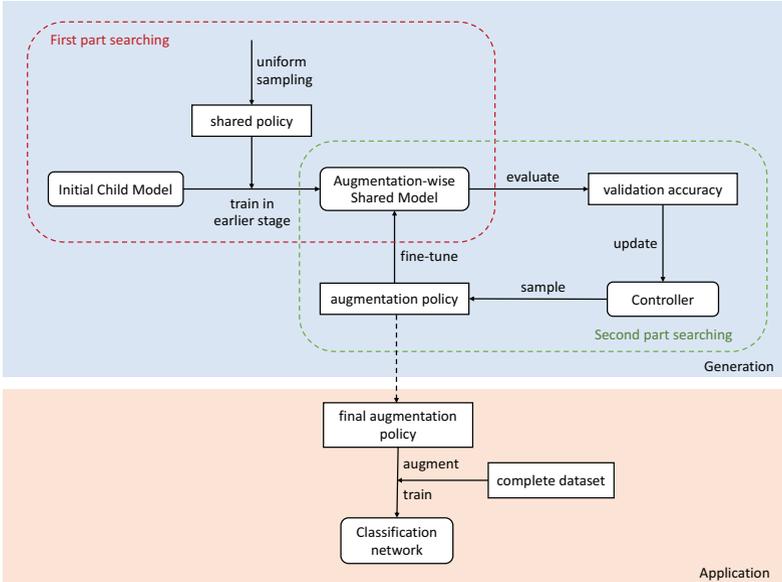}
  \caption{Augmentation-wise Weight Sharing workflow \cite{tian2020improving}. Upper/lower sections indicate the policy generation/application stage respectively.}
  \label{Fig: aws}
\end{figure}

Inspired by the idea of early stopping, the authors of AWS hypothesize that the benefit of DA is mainly shown in the later phase of training. This assumption is supported through empirical observations in \cite{tian2020improving}. Motivated by this observation, AWS proposes a new proxy task to test the sampled policies. In this proxy task, the original search stage is split into two parts. The AWS pipeline is displayed in Fig. \ref{Fig: aws}. In the early stage, the child model is trained using a fixed augmentation strategy, i.e. a shared policy. During this phase, the search controller will not sample policies or be updated. Only the child model used for policy evaluation will be trained for a certain number of epochs. The network weights obtained after the first part of training will be shared and reused during the later evaluation, so that AWS model does not need to repeat the full training for each of the sampled policies. The major challenge here is to select a representative shared policy for the initial stage. According to the findings in \cite{tian2020improving}, simple uniform sampling can work for most tasks. 

In the second part of searching, AWS samples augmentation policies via a controller, and updates the model according to an associated accuracy reward. The reward information is obtained from the shared model, instead of an untrained child model. Therefore, in order to evaluate the sampled policies, it is only necessary to resume training for a few epochs using these policies. Since the training of the child model in AWS is divided into two parts based on the different DA policies utilized, AWS is an augmentation-wise algorithm. The idea of weight sharing originates from NAS, where training from scratch is prohibitively expensive. This scheme substantially accelerates the overall evaluation procedure. The design of proxy tasks in AWS is flexible, so it can be combined with other search algorithms. Standard AWS follows a similar setting to the original AA \cite{cubuk2019autoaugment} applying Reinforcement Learning (RL) techniques. 

The major contribution of AWS is the effective period of the data augmentation technique. The empirical conclusion in \cite{tian2020improving} is that the DA policies mainly improve the model in the late training phase. This phenomenon reflects the greatest innovation in AWS, its unique augmentation-wise proxy task that substitutes the traditional evaluation procedure. By sharing the policy at the early phase of searching, the child model only needs to be pre-trained once. The selection of the shared augmentation policy in the first-part searching is done via a uniform sampling on the search space. The network weights are then re-used in the later application stage to evaluate each of the sampled augmentation policies. There is no need to conduct child model training from scratch thousands of times. Compared to the original AA \cite{cubuk2019autoaugment}, it is much more efficient to obtain reward signals in AWS through the use of weight-sharing strategies. The efficiency gains of AWS makes it have the potential to scale on even larger datasets. Moreover, according to \cite{tian2020improving}, the evaluation process in AWS is still reliable. This is because in the second part of searching, the child model will be fine-tuned by DA policies to reflect the strength of each of the policies. 

The disadvantages of AWS cannot be ignored however. Overall, there are excessive simplifications in AWS, aimed to increasing search efficiency. For example, sampled policies are evaluated by child models on the reduced data, and the early stage of training is substituted by shared model weights. Such settings however might lead to sub-optimal results. The final policy AWS model outputs may be more designed to the proxy task rather than the target dataset according to the findings in \cite{cubuk2020randaugment}. In terms of the search algorithm, AWS utilizes the same RL framework as in AA, bringing not much improvement, especially when compared with methods such as Fast AA \cite{lim2019fast} and PBA \cite{ho2019population}. Lastly, AWS is not open-source, which makes it less accessible for users.


\subsubsection{Greedy AutoAugment (GAA)}

To improve the search efficiency, Greedy AutoAugment (GAA) \cite{naghizadeh2020greedy, naghizadeh2021greedy} adapts a completely different algorithm. The GAA model applies a greedy search algorithm to exponentially reduce its complexity when sampling the next policy to be searched. From the experiments conducted in \cite{naghizadeh2020greedy}, the TFs learned by GAA are able to further enhance the generalization ability of the classification. Moreover, the greedy idea in GAA can be a reliable complement to other search approaches in AutoDA tasks. 

The policy model of GAA follows a similar setup in AA \cite{cubuk2019autoaugment}. A complete augmentation policy is comprised of $k$ sub-policies, each of which contains two consecutive TFs. Each TF is described by two essential hyper-parameters: probability and magnitude. The values of these two parameters are modeled following the same discretization as described earlier. There are $11$ values for probability parameter, ranging from $0$ to $1$ with uniform spacing, whilst the discrete values for magnitude are positive integers, range from $1$ to $10$. However, GAA employs a wider range of augmentation transformations. There are $20$ available image transformation functions in GAA that can be selected to form the DA policy, including $4$ extra operations compared to original AA. Assuming each augmentation policy contains $L$ image operations, where $L$ is a positive integer greater than $0$, then the search space can be defined as $(20\times11\times10)^L$. In this setup, the expansion of the search space is exponential to the value of $L$, which can be infeasible when using larger $L$ values.

To tackle this problem, the search space in GAA is re-formulated into a much simpler setup. It has been argued in \cite{naghizadeh2020greedy, naghizadeh2021greedy} that using separate probability parameters for each TF may not be necessary, especially when dealing with small amounts of data. Instead, all images in the original training set should be fully augmented till the enhancement of model performance becomes obvious. Therefore, GAA completely discards the probability hyper-parameter of each TF. Moreover, it adapts constant a value $1$ to represent the application probability of all available augmentation functions. Rather than optimising the probability hyper-parameter for each TF, GAA simply selects the TF that can give the best accuracy results. This is quite different from other AutoDA methods in how an augmentation policy is formed. By doing so, GAA successfully reduces the search space to $(20\times10)^L$. 

However, even with the reduced search space, the growth of such space is still exponential. GAA employs tree-based Breadth-First Search (BFS) to tackle this problem. In a standard BFS pipeline, the next search point is sampled in a greedy way, which means GAA will choose the TF that can bring the most performance gain to the child model. GAA only needs to evaluate the best image operation at each epoch, rather than all available TFs in the augmentation space. The greedy BFS changes the exponential growth of search space from $(20\times10)^L$ to a linear one. The size of final search space in GAA is $20 \times 10 \times k$, where $k$ is the number of sub-policies within one DA policy. By default, $k$ is set to $5$ in GAA for a fair comparison with AA. 

The search process of GAA is conducted as follows. Firstly, it goes through all possible TFs and their magnitude values, while the probabilities are all set to $1$. Then, each operation is scored with its respective accuracy value obtained from the training of child model. As discussed before, searching in GAA is conducted in a greedy manner. Accordingly, only the TF with the highest score/accuracy will be stored. It is then concatenated to the next operation. This search procedure will be repeated $k$ times to find the top $k$ best sub-policies. Each of these selected sub-policies will be concatenated with all previously learned TFs to form the final policy of GAA. 

The efficiency of GAA is substantially improved without a performance drop. From the experiments in \cite{naghizadeh2020greedy, naghizadeh2021greedy}, GAA requires $360$ times less computational cost than the original AA \cite{cubuk2019autoaugment} while still maintaining comparable model accuracy. Though the improvement in search speed is appealing, GAA has several limitations. The most significant disadvantage of GAA comes from its simplification of the search space. In GAA, the augmentation policy is formed by selecting TFs one after another, solely based on the performance of the selected operation. This setting assumes that each TF is independent even though they might affect each other in practice. As discussed in \cite{ratner2017learning}, certain image operations might have pretty different results depending on which other TFs are applied together. This conclusion is also supported by experiments in \cite{ho2019population}. However, due to the greedy nature of BFS, GAA only looks one step forward during the search, and hence can be easily trapped in local maximum that results in sub-optimal solutions. In addition, most of the hyper-parameters in GAA are mainly selected manually. There is also a lack of empirical evidence supporting such decisions. 


\subsubsection{Population-Based Augmentation (PBA)}

Population-Based Augment (PBA) \cite{jaderberg2017population} is one of the most widely accepted AutoDA approaches. Unlike GAA that models DA policies as independent TFs, PBA emphasizes the relationship between them. In PBA, the standard augmentation policy search problem is treated as a special hyper-parameter optimization, where the schedule of these parameters is stressed. The schedule here refers to the application sequence of TFs. The augmentation policy is not a fixed setting. Instead, it changes as training progresses. To accommodate additional sequential information, PBA leverages the Population Based Training (PBT) \cite{jaderberg2017population} technique. This algorithm optimizes the hyper-parameters along with the network weights simultaneously to achieve optimal performance. The final output of PBA is not a fixed configuration, but an application schedule of selected TFs when training the classification model. However, due to efficiency considerations, searching in PBA is still conducted on a simplified child model instead of the target network. PBA discards the trained child model as other AutoDA methods. The searched DA schedule is then adapted to the target training set to help train more complicated models. 

In order to directly compare with AA, PBA retains similar settings as much as possible. The same $15$ augmentation TFs are available in PBA except SamplePairing \cite{inoue2018data}. The discretion of policy hyper-parameters follows the same formulation, allowing $11$ values for magnitude and $10$ for probability. In PBA, a sub-policy still consists of two TFs, which are applied on one of the training batches. The policy modelling in PBA is motivated by the need for fair comparison with AA, rather than achieving optimal performance. Since the order of augmentation TFs in policy matters, PBA has an enormous search space even when compared with AA. For a single augmentation function, there are $(10\times11)^{15\times2} \approx 1.75\times10^{61}$ possibilities, much more than $(16\times10\times11)^{10} \approx 2.9\times10^{32}$ in AA \cite{cubuk2019autoaugment}. 

Despite having a larger search space, PBA demonstrates that searching for a schedule is considerably more efficient than enforcing a fixed regulation. This is due to several factors. In traditional AutoDA methods such as AA, evaluating the sampled policies is extremely time-consuming. Such process needs to be conducted via a full training of a child model, because data augmentation techniques primarily take effect in the later stage of model training. In order to estimate the effectiveness of a fixed policy, the child model has to be trained for a certain number of epochs till the model can actually benefit from the policy. However, it is totally different when testing a policy with an application schedule. If two newly sampled policies share the same prefix TFs, the evaluation algorithm can reuse the prior training weights for the evaluation of both policies. This is similar to weight-sharing idea in AWS \cite{tian2020improving} but it is more reliable. Moreover, it is also argued in \cite{ho2019population} that DA can provide better accuracy results when utilizing schedule information. Different types of augmentation TFs may be appropriate for different epochs during model training. It is a natural thought to choose the most suitable augmentation functions according to the training stage. 

\begin{figure}[h]
  \centering
  \includegraphics[width=0.7\linewidth]{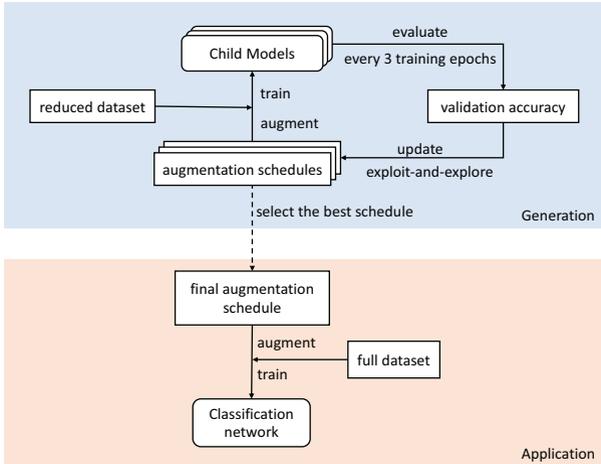}
  \caption{Population Based Augmentation workflow \cite{ho2019population}. Upper/lower sections indicate the policy generation/application stage respectively.}
  \label{Fig: pba}
\end{figure}

The basic workflow of PBA is displayed in Fig. \ref{Fig: pba}. To start, several child models, i.e. population, are initialized and trained concurrently. Each child model is responsible for evaluating a single augmentation policy. After $3$ epochs of training, PBA runs one epoch of Gradient Descent (GD) by testing all child models on the validation dataset. The validation accuracy estimates the performance of each policy used to train a child model respectively. After obtaining the performance rank of all child models and corresponding DA policies, PBA employs a classical “exploit-and-explore” procedure to update policies, where the lower ranked models copy the parameters of the higher ranker ones. Specifically, PBA uses Truncation Selection \cite{jaderberg2017population} for exploitation. For exploration, PBA randomly perturbs the parameter values during sampling and exploiting. PBA does not re-initialize the child model from scratch, which greatly reduces the computational cost. 

The improvement of PBA for search efficiency is substantial. It is approximately $1,000$ times faster than AA while still preserving similar accuracy. This is mainly due to the joint optimization of the child model and policy hyper-parameters. Even though the trained model is discarded after the search, re-using the network weights without repetitive training requires much less computational resources. The output of PBA is an application schedule for the augmentation policy. This can be represented as an augmentation function $f(t)$, with training epoch $t$ as a variable. The final output of PBA reports moderate probability values of all operations. This may be due to the random perturbation when updating the policies. The magnitude values of all augmentation TFs also share a pattern. In the early phase of training, the increase in magnitude values is rapid. As training progresses, all magnitudes will reach a stable state. explaining this phenomenon, the authors of PBA argue that an effective evaluation procedure should be conducted for at least a certain number of epochs till the DA policies fully function on the model \cite{ho2019population}. Moreover, the experimental results in \cite{ho2019population} also suggest that simple TFs might be more suitable in the initial stage, while more complicated DA operations can be applied later.


\subsubsection{Fast AutoAugment (Fast AA)}

Besides PBA, another widely accepted AutoDA is Fast AutoAugment (Fast AA) \cite{lim2019fast}. This method is motivated by Bayesian Data Augmentation \cite{tran2017bayesian} to solve the AutoDA problem. However, the objective of search phase in Fast AA is no longer focused on the highest model accuracy. Instead, Fast AA treats the augmented images as the data points from the original data distribution, which can best enhance the generalization ability of the model to be trained. Therefore, in Fast AA, the modified optimization objective focuses on minimizing the distribution distance between the original data and new data generated by DA policies. This is realized by adapting a density matching algorithm. This algorithm operates by matching the density of the original and the generated data, and hence completely eliminates the need for re-training the child model.

Similar to the aforementioned methods, Fast AA employs the same problem formulation as in AA \cite{cubuk2019autoaugment}, but uses continuous values instead. The two hyper-parameters of augmentation TF (probability and magnitude) have far more possibilities in Fast AA. However, in contrast to PBA \cite{ho2019population}, the application of Fast AA policies is random and without any sequential order. The final policy of Fast AA is a combination of $5$ sub-policies. Each sub-policy is a conjunction of $2$ TFs, following the same experimental setting of AA. The authors of Fast AA emphasize that the number of sub-policies can be further tuned during the search \cite{lim2019fast} because of the efficiency of the approach. 

\begin{figure}[h]
  \centering
  \includegraphics[width=\linewidth]{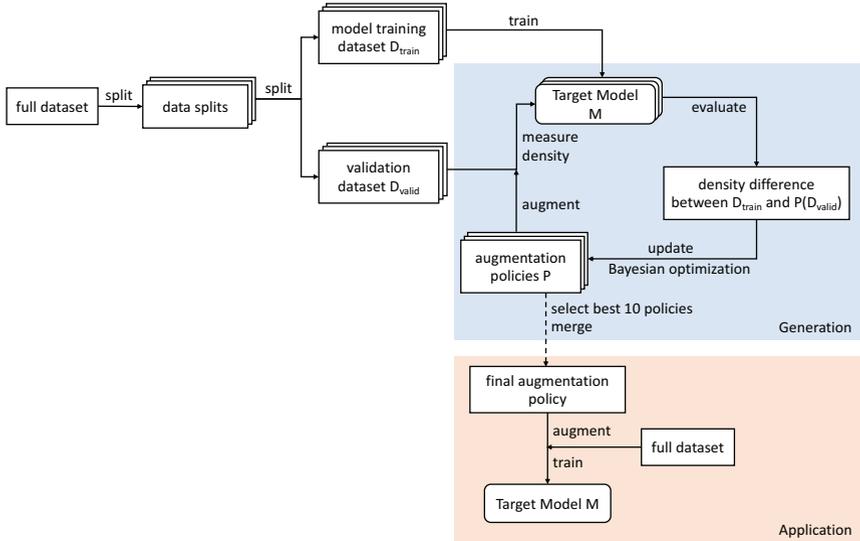}
  \caption{Fast AutoAugment workflow \cite{lim2019fast}. Upper/lower sections indicate the policy generation/application stage respectively.}
  \label{Fig: fast}
\end{figure}

Fast AA optimizes the augmentation policy during the process of density matching between training data pairs. Fig. \ref{Fig: fast} displays the general workflow of Fast AA. The target data is divided into two sets, namely the training data $D_{train}$ and validation data $D_{valid}$. At a high level, the goal of the search algorithm is to find a DA policy $P$ that can best match the density of $D_{train}$ and $P(D_{valid})$ augmented by $P$. Unfortunately, it is infeasible to directly compare data distributions for each of the sampled policies. Fast AA tackles this problem by approximating the data distribution via model predictions. Specifically, if the same model can achieve equally promising results over two datasets, it is reasonable to consider that two sets might share a similar distribution. 

In detail, Fast AA first splits the full data into several folds, each of which is assigned to a classification model $M$ and processed in parallel. Every data fold consists of two separated datasets $D_{train}$ and $D_{valid}$. Training data is used to pre-train the model $M$, while the validation data will be retained for the subsequent policy search. After pre-training, the Fast AA model starts to sample augmentation policies via a classical Bayes Optimization \cite{tran2017bayesian}. These policies are then evaluated based on the performance of the trained model $M$ on augmented data $P(D_{valid})$. Specifically, the policy that can generate data to maximize the performance of $M$ is desired. During the evaluation process in Fast AA, there is no training step involved at all. As such, it is a much more efficient and approach than other training-based evaluation methods. 

The experimental results in \cite{lim2019fast} show that this evaluation setup leads to superior speed, and provide competitive accuracy to AA on various data. Fast AA achieves promising results not only for direct policy search, but also the transfer of learned policies to new data. The evidence presented in \cite{lim2019fast} suggests that the transferability of Fast AA far exceeds the original AA. This is because Fast AA conducts the search directly on full datasets using target classifications, which minimizes the sub-optimality of learned policies. Therefore, Fast AA also has the potential to achieve better results when given more complex tasks. Making use of the efficiency of Fast AA model, the total number of sub-policies contained in an augmentation policy can be more than original setting. The experiments in \cite{lim2019fast} show an improved generalization performance with more sub-policies searched by Fast AA. Moreover, it is also possible for Fast AA to search augmentation policies for each class. By doing so, Fast AA also obtains a slightly improved performance. 

The biggest advantage of Fast AA lies in its efficiency. After introducing density matching into the evaluation process, Fast AA does not need to train the child model at all. As a result, Fast AA achieves a significantly faster search speed than the vanilla AA \cite{cubuk2019autoaugment}. Moreover, contrary to all previous AutoDA approaches \cite{cubuk2019autoaugment, ho2019population, tian2020improving}, Fast AA does not necessarily simplify the given task by searching on a proxy task. This avoids possible sub-optimiality, and hence guarantees the competitive performance of Fast AA. Another benefit of Fast AA is its continuous search space instead of discrete values, which finds more candidate policies during the search and potentially enhances the final result. Furthermore, the operations in Fast AA can be conducted in parallel, making it more practical to implement for ordinary users. 



\subsubsection{AutoAugment with Knowledge Distillation (AA-KD)}

Although AA-based algorithms have been a powerful augmentation method for many classification tasks, they are often sensitive to the choice of TFs. An inappropriate DA policy may even deteriorate the model performance. In traditional AutoDA methods \cite{cubuk2019autoaugment, lim2019fast, ho2019population}, usually there exists a trade-off between data diversity and label safety. Aggressive TFs can give more diverse training data that may better generalize the model, but they can potentially corrupt annotation information. In contrast, mild augmentation operations preserve the original image label, while the diversity of augmented data might be constrained and only lead to limited performance improvements. 

This has been investigated in AutoAugment with Knowledge Distillation (AA-KD) \cite{wei2020circumventing}. Examples in \cite{wei2020circumventing} show that some aggressive TFs can remove important semantics from the training data, which results in augment ambiguity. Augment ambiguity happens when the original label of a given image is no longer the most suitable annotation for the augmented data. Such phenomenon might confuse the classification model and result in performance drops. AA-KD approaches this problem by utilizing a Knowledge Distillation (KD) technique, which provides label correction after augmentation. 

The authors of KD-AA point out that aggressive transformations can still be helpful for model training, only under the premise that associated labels are adjusted accordingly. Using the original labels of all transformed images is not the best option. Therefore, each of the data samples should be treated differently based on the different effects of an augmentation operation. KD-AA leverages the idea of Knowledge Distillation to gain more diverse data as possible, while still providing accurate annotations. In KD-AA, a teacher model is used to generate complementary information for label corrections. Each transformed image is described by the original label as well as the teacher signal. The latter is also called a soft label, as the ground-truth label is slightly softened by the associated teacher signal. Such soft labels are then used during the training of student models in the policy application phase. The teacher-student framework in KD-AA is mainly designed to filter out biased or noisy annotations resulted from DA to further enhance the later model training. 

The major contribution of KD-AA is to recognise that KD is a useful technique to enhance traditional AutoDA methods. It is a complement to augmentation search algorithms rather than a complete AutoDA model. The effectiveness of KD-AA is demonstrated via experiments using AA \cite{cubuk2019autoaugment} and RandAugment \cite{cubuk2020randaugment} in \cite{wei2020circumventing}. With much larger magnitude values, AutoDA supported by KD produces consistent improvements in model accuracy. The authors of KD-AA also report the possibility of employing semi-supervised learning techniques with KD in AutoDA \cite{tarvainen2017mean, thornton2013auto, xie2020adversarial}, where the input data is mostly unlabelled. 


\subsubsection{Patch AutoAugment (PAA)}

Generally, AutoDA methods search for the augmentation policies at the image level. The same DA policy is used to transform the entire image. However, depending on different content within various regions of an image, the optimal TF may be different. Treating an image as a whole might ignore the difference in its internal regions, which constrains the diversity of augmented data \cite{gontijo2020affinity}. Moreover, overly aggressive augmentation functions may potentially modify or remove semantic features, causing safety concerns in terms of label preservation. To address the above-mentioned problem, a more fine-grained approach, Patch AutoAugment (PAA) \cite{lin2021local} has been proposed. PAA considers an image as a combination of several patches, and optimizes DA policies at patch level. To fully represent the inner relationship between each patch, PAA formulates AutoDA tasks as a Multi-Agent Reinforcement Learning (MARL) problem, where each agent handles a single patch and updates the final DA policies corporately. 

Similar to AA, the problem modelling in PAA follows the basic use of Reinforcement Learning (RL) model. In a standard RL framework, given a current state, the agent/controller samples a DA policy and receives the corresponding reward signal from the child model training. Then the PAA model updates the policy according to the reward and moves to the next state. The optimization objective of an agent in PAA is the maximization of reward to search for the best performing DA policy. Moreover, PAA employs the idea from MARL \cite{boutilier1996planning}, using multiple agents to search over a single image. Each agent handles a sub-region out of the original picture, and shares a global reward to cooperatively update the next augmentation strategy. 

In PAA, the augmentation policy model consists of a global state, local observations and actions. Firstly, an input image is divided into several non-overlapping patches of equal size. Each patch is controlled by one policy agent for the policy optimization. To accommodate the contextual relationship between other patches, a global state is shared among all agents, representing the semantics of the whole image. The global state is obtained by extracting the deep features of the entire input image through a standard CNN model. In PAA, the ResNet-18 network \cite{he2016deep} pre-trained on ImageNet data \cite{deng2009imagenet} is used. In addition to the global state, each agent also utilizes local information, i.e. observations, to update its own DA policy. Local observations are also described by the deep features of the associated patches. Unlike global information, this information is unavailable to other agents. During the search, each of the agents update its policy based on both global and local information. The action of the policy model in PAA represents standard TF techniques, controlled by the probability and magnitude. There are $15$ operations defined in PAA. Similar to PBA \cite{ho2019population}, PAA outputs an application schedule of TFs instead of fixed policies. However, since the search in PAA is conducted on a child model for a limited number of epochs, such schedule needs to be linearly scaled up when applied to target networks in the second stage. 

The optimal strength of PAA is its patch-based policy search. Through the use of Grad-CAM \cite{selvaraju2017grad}, the importance of each region in the image is clearly shown in \cite{lin2021local}. This further shows that the optimal augmentation strategy for each patch can vary depending on the different semantics. It is therefore reasonable to perform the augmentation search at a more fine-grained level. Different patches might prefer totally different TFs during the training. For example, regions that contain important features may prefer mild TFs, which can better preserve the semantics. However for less important patches that do not include objects of interest, aggressive operations with larger magnitude values might provide a higher level of variety in the augmented data. Overall, a fine-grained PAA can not only provide sufficient variety for the proposed policies, but also further enhance model performance.


\subsection{Gradient-based Optimization}


\subsubsection{Faster AutoAugment (Faster AA)}

From the perspective of hyper-parameter optimization, all of the aforementioned two-stage AutoDA methods do not directly optimize augmentation policies via gradients. This is because the augmentation operations are usually not differentiable with respect to the hyper-parameters of policy probability and magnitude \cite{hataya2020faster}. As a result, it is often tricky to obtain the gradient information of validation accuracy with regard to policy hyper-parameters \cite{lin2019online}. However, several works have proposed to approximate the hyper-parameters as probability distributions, and relax such distributions in a way that a DA policy can be optimized based on gradients \cite{hataya2020faster, lin2019online, gudovskiy2021autodo, li2020dada}. Faster AutoAugment (Faster AA) \cite{hataya2020faster} is the only two-stage approach that is based on gradient optimization.

Faster AA also employs a similar policy model as in previous works \cite{cubuk2019autoaugment, ho2019population, lim2019fast}. In Faster AA, a DA policy consists of several sub-policies, where a single sub-policy contains $2$ consecutive augmentation TFs. Each TF is described by two hyper-parameters: the probability and magnitude. Operations used in Faster AA are the same $16$ image TFs from the original AA work, including $14$ basic image transformations functions implemented in PIL library and $2$ extra augmentation algorithms, i.e. Cutout \cite{devries2017improved} and SamplePairing \cite{inoue2018data}. Since a DA policy is often non-differentiable based on its hyper-parameters, traditional AutoDA models have to conduct a full training on child model to evaluate a policy. Even after the discretization of policy hyper-parameters, this formulation still requires exorbitant computational resources \cite{cubuk2019autoaugment, tian2020improving, ho2019population}. To address this challenge, Faster AA approximates the original search space into a differentiable setting and directly optimizes the gradient, which significantly reduces the search cost.

Inspired by the bi-level optimization in OHL-AA \cite{lin2019online}, Faster AA adapts a differentiable framework for DA policy search, which substantially accelerates the search. The key modification in Faster AA is the approximation of policy gradient via a straight-through estimator \cite{bengio2013estimating, oord2017neural} inspired by DARTS \cite{williams1992simple}. The success of DARTS in NAS fields makes it suitable to AutoDA tasks. In Faster AA, the distribution of augmentation hyper-parameters are approximated as Relaxed Bernoulli distribution \cite{jang2016categorical}. After the relaxation of search space, each TF within an augmentation policy can be differentiable with regard to the probability and magnitude hyper-parameters \cite{bengio2013estimating, oord2017neural}. This makes it easier to calculate their gradients in Faster AA. Using gradient estimation techniques, Faster AA can directly optimise DA policies based on gradient. This makes DA policy optimization end-to-end differentiable, and thus provides much more control over the entire process especially when compared to previous black-box optimizations, such as Reinforcement Learning in AA \cite{cubuk2019autoaugment}.   

\begin{figure}[h] 
  \centering
  \includegraphics[width=\linewidth]{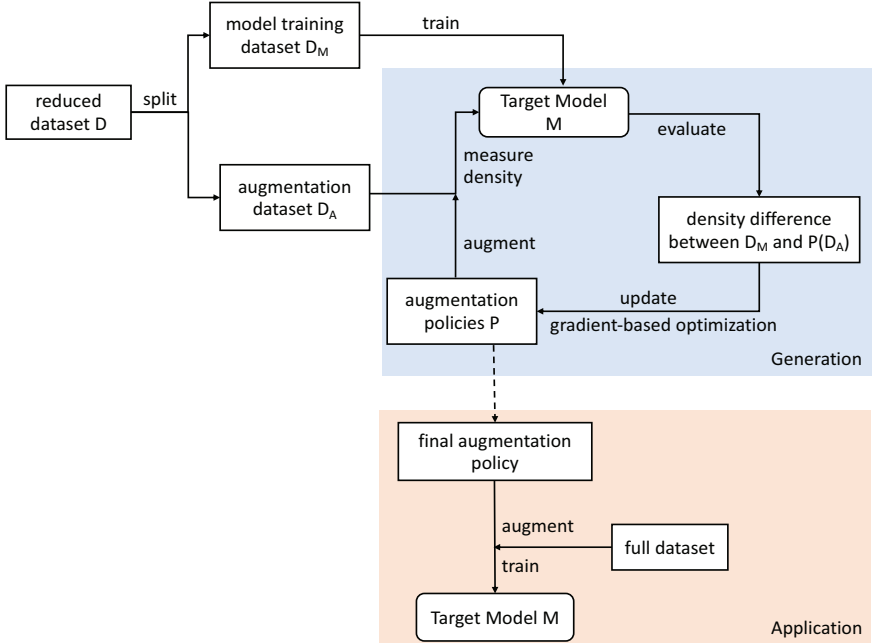}
  \caption{Faster AutoAugment workflow \cite{hataya2020faster}. Upper/lower sections indicate the policy generation/application stage respectively.}
  \label{Fig: faster}
\end{figure}

To further reduce the search cost, inspired by Fast AA \cite{lim2019fast}, Faster AA also applies a density matching technique during policy evaluation. The objective of optimization is to minimize the distance between data distributions of the original and the augmented data. Faster AA also employs an adversarial framework to help the policy sampling. Due to the use of Density Matching, the overall workflow of Faster AA is similar to Fast AA as displayed in Fig. \ref{Fig: faster}. However, Faster AA uses a reduced dataset $D$ as its input to further improve the search efficiency. Input data $D$ is firstly split into two sets: a training set $D_M$ to prepare the evaluation model and an augmentation set $D_A$ for policy searching. The pre-trained model $M$ is then used to estimate the performance of sampled DA policies $P$ in the first stage of Faster AA. The Faster AA model examines the search space in a gradient-based manner to identify the optimal DA policies. After searching, the final policy will be applied to augmented the full dataset and train the classification model $M$. During both the searching and training phase, Faster AA trains the same target model $M$ to provide more reliable evaluation. 

Faster AA is the first two-stage AutoDA model that resorts to gradient approximation to achieve faster searches than other state-of-the-art algorithms, such as Fast AA \cite{lim2019fast} and PBA \cite{ho2019population}. Importantly, it introduces straight-through estimators \cite{bengio2013estimating} to approximate the gradient of non-differentiable AutoDA task. By relaxing the original distributions of policy hyper-parameters, Faster AA can directly back-propagate the augmentation process and optimize DA policies based on the gradient. The black-box optimization used in traditional two-stage AutoDA models is therefore transformed into a more transparent and controlled process to significantly improve the search speed. In addition, Faster AA follows the density matching idea of Fast AA \cite{lim2019fast}, completely removing the repetitive training of the model in the first stage. Instead, policy evaluation is conducted by minimizing the distribution distance between the original and the augmented data. As a result, Faster AA can substantially reduce the required computational resources compared to AA \cite{cubuk2019autoaugment}. The experimental results in \cite{hataya2020faster} shows the competitive performance of Faster AA compared to other AutoDA methods, in terms of both search efficiency and final accuracy.



\subsection{Search-free}

\subsubsection{RandAugment (RA)}

RandAugment (RA) \cite{cubuk2020randaugment} is the first search-free scheme in the AutoDA field. To reduce the cost of the search phase, the parameter space in RA is significantly smaller, defined by only two hyper-parameters. This reduced space allows RA to learn DA policies directly from the full dataset without resorting to a separate proxy task. In fact, the simplification of policy hyper-parameters in RA is so dramatic, that a simple grid search is sufficient to output an effective DA policy. Therefore, the policy generation stage in RA is quite different from the classical search scheme in other approaches, where the latter usually involves selective sampling and expensive evaluation. According to \cite{cubuk2020randaugment}, it is possible to apply more advanced sampling methods instead of a naive grid search, which may further reduce the computational cost. Therefore, RA can be recognised as a search-free AutoDA model. Additionally, RA is also able to optimize DA policies based on different sizes of classification models and training data. The experimental results in \cite{cubuk2020randaugment} also show that RA can produce competitive accuracy result when compared with other search-based AutoDA approaches. 

The primary goal of RA is to reduce the complexity caused by the separate policy search stage in the earlier two-stage AutoDA methods. To do so, RA eliminates the need for expensive policy searches by greatly simplifying the search problem. The entire search phase in traditional AutoDA method is removed in RA out of efficiency considerations. This is because most of the computational workload comes from the first stage when the model repetitively samples DA policies and evaluates them. It is also a complicated bi-level optimization problem to conduct the policy search and network training simultaneously. Another downside of prior search-based methods lies on the proxy task used during searching. In the proxy task, AutoDA models search on a reduced sub-set of the original training data, and evaluate sampled augmentation policies using a simpler network. Both simplifications are applied in order to decrease the search cost. A major premise of this framework is that such a proxy task can reflect some core features of the target task, so that the final policy is also the optimal augmentation scheme for the full training data. While the DA policy found through the proxy task is able to produce promising performance \cite{cubuk2019autoaugment, ho2019population, lim2019fast, zoph2020learning}, it is likely to be a sub-optimal result \cite{cubuk2020randaugment}. According to \cite{cubuk2020randaugment}, the optimal strength of an augmentation policy depends on the size of both the training set and network. Therefore, searching for DA policies on a proxy task can only produce results suitable to solve the proxy task instead of the target task, which leads to sub-optimal solutions.

To avoid sub-optimal results, AutoDA models need to directly search for DA policies over the full training set. However, this is usually computationally infeasible in practice as the traditional search space in AA \cite{cubuk2019autoaugment} is extremely large. In order to mitigate such efficiency issues, RA substantially reduces the number of  hyper-parameters to optimise. In RA, the reduction in the size of search space is tackled in two ways, including the simplification on the existing formulation and the proposal of new parameters. In prior methods \cite{cubuk2019autoaugment, lim2019fast, ho2019population}, each TF in an augmentation policy is controlled by two hyper-parameters, probability and magnitude. While in RA, all image operations are selected with uniform probability, which depends entirely on the total number of available TFs in the search space. For instance, given $K$ different TFs in RA, the probability of applying each operation is $\frac{1}{K}$. 

To further reduce the search space, RA simplifies the magnitude hyper-parameter as well. The value range of the magnitude hyper-parameter follows the same setting as in original AA \cite{cubuk2019autoaugment}, with $11$ discrete values in total, ranging from $0$ to $10$. In previous AutoDA models, the scale of each transformation function is also specified by its respective magnitude. However, after examining changes in each operation magnitude during searching \cite{ho2019population}, the authors of RA point out that all magnitude values follow a similar schedule over time. Therefore, RA postulates that it may be sufficient to use a shared magnitude hyper-parameter $M$ for all TFs. As a result, in RA, all image operations within DA policies share the same probability and magnitude hyper-parameters, which significantly reduces the parameter space. Besides reformulating the search space, RA also proposes a new free parameter to improve the performance gain, namely the number of TFs $N$ within one augmentation policy. $N$ is predominantly manually decided in most popular AutoDA methods \cite{cubuk2019autoaugment, ho2019population, lim2019fast} due to limited computational resources. While in RA, automating the search of $N$ becomes feasible because of the extremely reduced parameter space. Optimizing the TF number $N$ can eliminate human bias and further improve performance. 

After the re-parameterization of parameter space, there are only two hyper-parameters to optimize in RA: the number of TFs $N$ to form a complete augmentation policy, and the global magnitude value $M$ to control all TFs. Both hyper-parameters can be easily interpreted by humans so that larger values of $N$ and $M$ indicate more aggressive augmentation strategies, while smaller values represent more conservative schemes. After RA has reformulated the entire search problem, various advanced algorithms can be applied to perform standard hyper-parameter optimization \cite{snoek2012practical}. However, since the final search space in RA is extremely small, the authors of RA suggest that a simple grid search can yield sufficient performance gains, which is supported by experiment \cite{cubuk2020randaugment}. 

RA makes several noteworthy contributions to the AutoDA task. Via re-parameterization of standard AutoDA problem, RA employs a reduced search space, which is only controlled by two hyper-parameters, $N$ and $M$. $N$ indicates how many TFs are contained within a single DA policy, and $M$ refers to the uniform distortion parameter for all image operations. In RA, the optimization of $N$ and $M$ is achieved by naive grid search. This feature allows RA to easily scale to larger datasets and deeper models without significantly increasing the search cost. Moreover, RA shows promising performance on various datasets, matching or even outperforming previous AutoDA models including AA \cite{cubuk2019autoaugment}, Fast AA \cite{lim2019fast} and PBA \cite{ho2019population}. This finding demonstrates the limitations of prior approaches based on proxy tasks. The experimental results in \cite{cubuk2020randaugment} are also in agreement with this finding, which shows that the optimal DA policy depends on the size of training data and discriminative network. Transferring a DA policy learned from a simplified proxy task can lead to performance degradation. After removing the expensive search phase, RA avoids the sub-optimality of learned DA policies through direct searches on the target dataset and classification model. Finally, the results in \cite{cubuk2020randaugment} reveals the relationship between augmentation policy and the size of dataset and model. Most existing AutoDA methods optimize DA policies using reduced data and smaller models to accelerate the search \cite{cubuk2019autoaugment, lim2019fast, ho2019population}, however this leads to sub-optimal performance. In practical applications, searching a full dataset can be computationally infeasible. Therefore, the findings in \cite{cubuk2020randaugment} have stimulated future innovations, aimed at balancing effectiveness and efficiency.



\subsubsection{UniformAugment (UA)}

UniformAugment (UA) \cite{lingchen2020uniformaugment} is another search-free method that also significantly reduces the parameter space. Unlike RA which employs a grid search to tune its augmentation parameters, UA completely eliminates the need for hyper-parameter optimization. Instead, UA restricts the range of values over which the policy hyper-parameters can be sampled, so that all DA policies falling into this range can preserve the original label of most of the data. Such a range is defined as an approximately invariant augmentation space in UA. A simple uniform sampling from the invariant space can produce effective DA policies, and eventually lead to sufficient performance gains. As a result, UA greatly surpasses all existing AutoDA models in terms of efficiency. The efficacy of UA is also demonstrated by extensive experiments \cite{cubuk2020randaugment}. Using the same $15$ augmentation TFs that are implemented in AA \cite{cubuk2019autoaugment} and other approaches \cite{lim2019fast, ho2019population, cubuk2020randaugment}, UA achieves comparable improvements in model accuracy. Furthermore, due to the removal of the search phase, UA is by far the most scalable AutoDA method, and can be easily applied to different tasks in the real world.

The key concept in UA is the introduction of invariant augmentation space. In \cite{lingchen2020uniformaugment}, an approximately invariant space is defined as a selected value range for the policy hyper-parameters. Each DA policy sampled from such a space is able to retain the representative features of the original data after transformation. In other words, most of the augmented data can still remain within the distribution of the original training set, without change of label information. From the perspective of Group Theory \cite{chen2020group}, when given such an invariant augmentation space, further optimizing policy hyper-parameters within this space can only yield limited performance gains, and is therefore unnecessary in practice. In that case, a naive random sampling approach might also lead to effective strategies, thus avoiding expensive computing cost. The experiments in \cite{lingchen2020uniformaugment} demonstrates the promising performance of UA, supporting the invariance assumption. However, UA is based on the premise that an invariant augmentation space is already known. While in UA, the invariant range is actually manually decided via empirical evidence from prior works \cite{cubuk2019autoaugment, lim2019fast, ho2019population, cubuk2020randaugment}. 
 
Similar to RA \cite{cubuk2020randaugment}, UA also explores the influence of two hyper-parameters $M$ and $N$, where $M$ refers to the operation magnitude and $N$ is the total number of TFs in a given policy. According to empirical evidence and theoretical analysis \cite{perez2017effectiveness, deng2009imagenet}, a good augmentation policy should be approximately invariant in order to generate in-distribution data, while being able to maximize data variety at the same time \cite{lingchen2020uniformaugment}. Usually, the generalizability of a classification network can be improved if the model is trained on more diverse data. This assumption emphasizes the importance of the value range for the magnitude $M$. Constraining $M$ within a narrow range can result in limited data diversity, whereas sampling policies from a wide $M$ range may produce overly aggressive TFs, which can remove original label information. Both results are considered sub-optimal, which suggests that there is a trade-off between diversity and correctness of learned DA policies. 

A similar trade-off can be found during the experiments on hyper-parameter $N$. Usually, optimizing $N$ is impractical in most prior AutoDA methods due to limited computational resources. However, it is possible to examine various $N$ values in search-free models such as RA \cite{cubuk2020randaugment} and UA. From the results in \cite{lingchen2020uniformaugment}, a smaller $N$ value usually indicates safer augmentation policies with less TFs applied on the image data. The transformed data tend to be less diverse. In contrast, a larger $N$ value might impose stronger DA operations on the training data and hence has the possibility of corrupting the original labels. In order to obtain the optimal strength of data augmentation, AutoDA models need to balance between effectiveness and safety feature when choosing $N$. After systematic experiments on $N$ values in \cite{lingchen2020uniformaugment}, $N$ in UA is set to $2$. This is sufficient to effectively improve model performance, while maintaining the same data distribution after augmentation. Moreover, $N = 2$ is also in line with the original proposal of AA \cite{cubuk2019autoaugment}. 

The contribution of UA has been revolutionary in the field of AutoDA. It not only proposes an effective automated DA scheme, but also substantially surpasses all existing approaches in terms of efficiency. More importantly, the hypothesis of augmentation invariance in \cite{lingchen2020uniformaugment} challenges the central premise of the AutoDA field. Most prior AutoDA models are motivated by the need for automatically searching for optimal augmentation hyper-parameters on given datasets, replacing biased and sub-optimal manual design. However, the necessity of such searching is questioned in \cite{lingchen2020uniformaugment}. Authors of UA propose the definition of an invariant augmentation space, in such a way that optimising DA hyper-parameters within that space is not necessary. This assumption is theoretically supported by group theory in \cite{chen2020group}. The comparable performance of UA also provides empirical evidence for the validity of the invariance hypothesis in data augmentation. 

However, how to decide an approximately invariant space for augmentation policy remains an open question. The efficacy of UA is mainly based on the application of domain knowledge, adapted from previous works \cite{cubuk2019autoaugment, lim2019fast, ho2019population, cubuk2020randaugment}. However, in the real-world scenarios, a given task and domain might be unknown. In addition, the pre-defined space of UA is also not guaranteed to be invariant, due to the lack of theoretical supports for its selection strategy. Even though UA yields positive results in empirical research, it is very likely that the final performance of a classification model can be further improved through the use of a more reliable policy space. According to \cite{cubuk2020randaugment}, it is important to develop a systematic methodology to determine an invariant policy space when given a specific dataset. Once such a range is decided, it is no longer necessary to perform an expensive search for DA hyper-parameters. Any augmentation strategy sampled from a invariant space should be effective enough for the given task to produce promising performance gains.



\section{One-stage Approaches}
\label{Sec: one}

After the pioneering AutoDA works such as AutoAugment (AA) \cite{cubuk2019autoaugment}, it is intuitive to approach AutoDA problems from a two-stage perspective. In the first policy generation phase, AutoDA models generate the optimal DA policy for a given dataset. In the second stage, the learned policy is then applied on the training set for model training. Optimization of policy hyper-parameters and network weights are performed in strict order. However, the separate generation stage in two-stage approaches results in additional computational complexity, which is also the major reason for their efficiency issues. For example, the original AA \cite{cubuk2019autoaugment} requires thousands of GPU hours to learn an effective augmentation policy better than the baseline. 

To improve the efficiency of AutoDA models, one idea is to perform the policy generation and application simultaneously, thus forgoing the extra computation in two stages. Despite inefficiency of the two-stage methods, it is inherently impractical to merge two stages and optimize the DA policy along with the classification model. This is because tuning policy hyper-parameters based on model performance is not a differentiable optimization problem. In other words, the gradients of augmentation hyper-parameters cannot be directly calculated nor optimized, thus precluding the possibility of joint optimization. However, with advances of gradient approximation techniques from Hyper-parameter Optimization (HPO) field, it is feasible to relax the original distribution and estimate policy gradients, allowing for one-stage AutoDA models.

This section reviews existing one-stage AutoDA methods including Online Hyper-parameter Learning AutoAugment (OHL-AA) \cite{lin2019online}, Adversarial AutoAugment (AAA) \cite{zhang2019adversarial}, Differentiable Automatic Data Augmentation (DADA) \cite{li2020dada} and Automated Dataset Optimization (AutoDO) \cite{gudovskiy2021autodo}. All of these approaches are based on gradient approximation through the use of differentiable frameworks. We discuss the formulation of optimization problem in each method. In particular, we focus on gradient approximation, which is the core technique in one-stage models. Lastly, we discuss the main contributions and limitations of each method.

\subsection{Gradient-based Optimization}



\subsubsection{Online Hyper-parameter Learning AutoAugment (OHL-AA)}

Before the proposal of Online Hyper-parameter Learning AutoAugment (OHL-AA) \cite{lin2019online}, the majority of the works in automated DA optimization followed the basic two-stage procedure \cite{cubuk2019autoaugment, lim2019fast, ho2019population}. Despite promising performance, most two-stage methods have serious bottlenecks in search time and cost. The authors of OHL-AA argue that the major cause of efficiency issues is the offline search, i.e. policy searching is performed independently of the final model training. In contrast, OHL-AA model applies an online scheme, where the policy hyper-parameters and network weights are optimized jointly in a single pass. In OHL-AA, policy hyper-parameters are formulated as probability distributions. Additionally, the gradients of DA policy are approximated via the use of the REINFORCE estimator \cite{williams1992simple}, which allows for direct optimization during model training. The final outputs of OHL-AA not only include the optimal DA policy for given task, but also contain the completely trained classification network. By combining the searching and training stages, the OHL-AA model reduces additional computational costs resulting from two stages, while still maintaining comparable performance.

Following the problem formulation in AA \cite{cubuk2019autoaugment}, OHL-AA also approaches AutoDA problem from the perspective of hyper-parameter optimization, but using a different optimization framework. Specifically, the DA policy in OHL-AA is sampled from a parameterized probability distribution, whose parameters are regarded as DA hyper-parameters that are optimized along with network weights. The joint optimization is achieved via a bi-level framework \cite{colson2007overview}. There are two layers of optimization in this bi-level setting. The inner objective is the training of classification model, while the outer objective is the optimization of the policy hyper-parameters through the use of REINFORCE gradient approximator \cite{williams1992simple}. Due to this bi-level optimization, OHL-AA is also acknowledged as an online AutoDA model, where the DA policy is updated together with the classification network. By optimizing the DA policy and task model simultaneously, OHL-AA completely discards the searching on small proxy tasks. Unlike previous two-stage methods \cite{cubuk2019autoaugment, ho2019population}, policy optimization in OHL-AA no longer requires thousands of evaluations, e.g. training of surrogate models. OHL-AA can directly optimize the DA policy through classical gradient descent algorithm, which substantially improves the search efficiency. 

\begin{figure}[h]
  \centering
  \includegraphics[width=\linewidth]{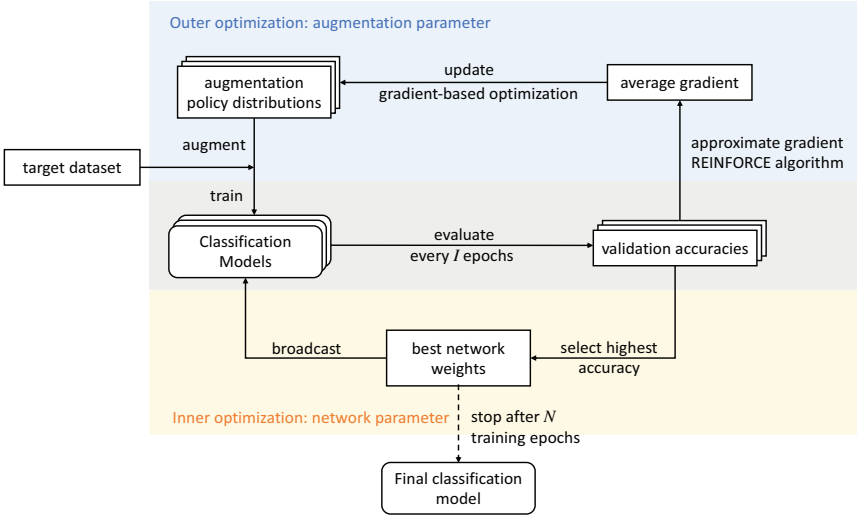}
  \caption{Online Hyper-parameter Learning AutoAugment (OHL-AA) workflow \cite{lin2019online}.}
  \label{Fig: ohl}
\end{figure}

Inspired by \cite{franceschi2017forward}, policy hyper-parameters in OHL-AA are updated in a forward manner. Specifically, the network weights after a certain number $I$ of epochs of inner optimization are forwarded to the outer optimization for an update at the outer level. In other words, the inner objective is updated for $I$ steps between two adjacent updates at the outer level. The overall workflow of OHL-AA is illustrated in Fig. \ref{Fig: ohl}. The bi-level optimization problem in OHL-AA is represented by two overlapping loops. In the inner loop, a group of the same classification models are trained in parallel using different DA policies, each of which is designed to evaluate the efficacy of the associated augmentation policy. After $I$ training epochs, all models are evaluated on the validation set. Among these models, the one with the highest validation accuracy is selected and broadcast to other models to synchronize network weights. 

Outer optimization is performed along with the inner procedures. In the outer loop, after the same $I$ number of inner updates, accuracy values are used to optimize the probability distributions of the DA policy. To be specific, after obtaining the validation accuracies of models, OHL-AA first calculates the average gradient of them using the distribution hyper-parameters (via the REINFORCE algorithm \cite{williams1992simple}). Such gradients are then used to update the policy distributions as a one step of gradient ascent. After the update, new augmentation policies are sampled from the updated distributions, and then used to train a group of synchronized networks. The whole process continues iteratively until the network or policy distribution finally converges. Overall, OHL-AA aims to find the optimal policy distribution that can generate the best DA policy. This framework drastically reduces the search cost, as the policy hyper-parameters are updated using only $I$ steps of model optimization instead of a complete training. 

The biggest contribution of OHL-AA is the proposal of the bi-level framework. It is also considered to be the first one-stage AutoDA model that optimizes both the DA policy and the network weights in a single pass. The removal of repetitive model training and proxy tasks significantly reduces the overall search cost. According to the results in \cite{lin2019online}, OHL-AA is $60\times$ faster than the original AA \cite{cubuk2019autoaugment} on CIFAR-10 and $24\times$ faster on ImageNet data, while still maintaining comparable model performance. In addition, the probability distribution of the DA policy in OHL-AA provides a feasible differentiation method for estimating policy gradients of AutoDA problems, which stimulates innovations in later one-stage approaches, such as DADA \cite{li2020dada} and AutoDO \cite{gudovskiy2021autodo}.



\subsubsection{Adversarial AutoAugment (AAA)}
Adversarial AutoAugment (AAA) \cite{zhang2019adversarial} is another one-stage AutoDA model that simultaneously optimizes the target network and augmentation policy. In additional to gradient approximation of DA policy, AAA innovatively employs adversarial concepts of GANs, leading to a more computationally efficient AutoDA approach. The ultimate goal of the AAA method is to best train the target classification model, rather than searching for the optimal DA policy. Similar to OHL-AA \cite{lin2019online}, training and searching in AAA are conducted in an online way, where the augmentation policy is dynamically updated along the training of discriminative model. Such procedures in AAA avoid the need for re-training the classification model, which significantly decreases the computational cost.

AAA preserves the standard formulation of the AutoDA problem in AA \cite{cubuk2019autoaugment}. In AAA, a complete augmentation policy for full dataset contains $5$ sub-policies. Each sub-policy is applied on one data batch before training the target model. A sub-policy is composed of two separate augmentation transformation functions, each of which is controlled by two hyper-parameters, i.e. probability and magnitude. To obtain better performance and easily compare with the original AA \cite{cubuk2019autoaugment}, AAA precludes the probability factor of augmentation operations during training. According to \cite{zhang2019adversarial}, such hyper-parameter requires a certain number of training epochs to take effect. This is effective for offline frameworks such as AA \cite{cubuk2019autoaugment}, because policy models are updated based on the result of full training on a child model. However, in an online AAA model, DA policy is dynamically evolved along with the training of target networks. The number of training epochs for each update is not sufficient to fully demonstrate the randomness of image operations, which may constrain the optimal strength of reward signal. 

In AAA, network training and DA policy generation is performed simultaneously. Different from standard two-stage approaches, the augmentation policy is dynamically updated rather than fixed during model training. As with other one-stage AutoDA methods, AAA also needs to perform gradient approximation on DA hyper-parameters to support joint optimization in a non-differentiable framework \cite{wang2017fast, peng2018jointly}. Specifically, the REINFORCE algorithm \cite{williams1992simple} is applied in AAA to estimate the policy gradient. The overall framework of AAA follows a standard GAN structure, consisting of a policy model as well as a target model. The training of these two models is formulated as a min-max game in an adversarial way. The policy model here is regarded as an adversary. During the training, the target network aims to minimize the training loss over the input data, while the objective of the policy network is to maximize the training loss of the target model by generating adversarial DA policies. These adversarial policies force the target model to learn from harder data samples and thus substantially improve its generalizability. When updating the policy network, the reward signal used comes from the training losses of target network after normalization. These loss values are associated with different augmentation strategies to indicate the efficacy of DA policies respectively. 

The major motivation for proposing AAA is the limited randomness of traditional policy search. Although the enormous search space in most AA-based approaches allows for a large variety of policy candidates \cite{cubuk2019autoaugment, lim2019fast}, fixing the sampled policy during the entire model training often leads to an inevitable overfitting problem \cite{ho2019population}. To tackle this issue, AAA chooses to use a dynamic augmentation policy, which is updated based on the state of the target model during training. The concept of dynamic DA policy was first proposed in PBA \cite{ho2019population}, where the application schedule of TFs was especially emphasized by sharing TF prefixes. While in AAA, the stochasticity of the policy search was further enhanced, as the entire DA policy was updated along with model training, rather than just a subset of TFs within one policy. The increased randomness in policy sampling provides the augmented data with more diversity, and thus can better train the target model.   

Another motivation of AAA is the efficiency problem existing in most AutoDA methods. The first AutoDA model AA \cite{cubuk2019autoaugment} was largely criticized due to its excessive training time. Later works such as PBA \cite{ho2019population} manage to accelerate the whole process by trading time with space. Although the overall search time is substantially decreased, training a large population of child models simultaneously still requires significant computational resources. AAA, however, is considered to be computation-efficient and resource-friendly. During the search in AAA, the target model only needs to be trained once. By reusing the prior computation in training, policy networks are updated based on the intermediate state of target models instead of the final result. By the end of training, the target network is supposed to be optimized via combating adversarial policies. Due to the reduced computational cost and time overheads, AAA can directly perform searching on the full data using the target network. A direct search not only guarantees the effectiveness of AAA, but also eliminates the potential sub-optimality that may result from employing proxy tasks. 

The most significant innovation in AAA is its adversarial framework. In fact, adversarial learning is not the first time it has been utilized in AutoDA problems. The earliest TANDA approach \cite{ratner2017learning} also employed standard GAN structures, which used policy models as generators to sample DA policies, while another discriminative network was used to identify augmented samples out of the original data for policy evaluation. On the contrary, AAA uses policy model as an adversary against the training of the target model. The policy model in AAA produces aggressive DA policies that maximize the training loss. The data transformed by these policies are often more distorted, making it more difficult for target models to distinguish, which in turn allows the target model to learn more robust features via adversarial learning. The final goal of AAA is not just finding the optimal policy on given dataset. Instead, AAA places more emphasis on the final result of target model training. By feeding deformed examples, AAA trains the classification network, allowing it to be more resilient to a variety of data points, thereby greatly enhancing its ability of generalization. 

In addition, AAA also outperforms previous AutoDA methods in terms of evaluation. AAA directly evaluates the performance of target model via training loss, while in methods such as TANDA \cite{ratner2017learning} or Fast AA \cite{lim2019fast}, the effectiveness of DA policies is estimated using the similarity of the augmented data to the original data. Generally, it is considered to be an effective policy if transformed pictures resemble the original samples. Such approximation might result in potential performance degradation due to the lack of variety in the learned policies. However, this issue can be substantially mitigated in AAA as the training loss is used as an intuitive criteria to evaluate the target model. The experimental results in \cite{xie2020adversarial} also show that classification networks trained in AAA have higher accuracy than previous methods.



\subsubsection{Differentiable Automatic Data Augmentation (DADA)}

Motivated by the development of differentiable NAS \cite{dong2019searching, liu2018darts, simonyan2014very}, a number of one-stage AutoDA approaches have been proposed following OHL-AA \cite{lin2019online}. Differentiable Automatic Data Augmentation (DADA) is another effective method that relies on gradient approximation to optimize model weights and DA policy at the same time. The basic gradient estimator utilized in DADA is the Gumbel-Softmax gradient estimator \cite{jang2016categorical}. Additionally, DADA also proposes a new estimator named RELAX, which is designed to solve the imbalance problem of training data. By combining two gradient approximators, DADA suggests an effective and efficient DA policy learning, which is more robust to biased or noisy data. 

The formulation of the DA policy in DADA follows the standard setting of AA \cite{cubuk2019autoaugment}. A complete augmentation policy is comprised of $25$ sub-policies, each of which will be used to augment one data batch of training set. One sub-policy contains two TFs, which are described by two hyper-parameters (probability and magnitude). Similar to AAA \cite{zhang2019adversarial}, DADA also uses the probability distributions to encode augmentation TFs when sampling DA policies. The sub-policy is sampled from the categorical distribution, and the hyper-parameters of each transformation are approximated as Bernoulli distributions. After the re-parameterization of the search space, the AutoDA task is formulated as a Monte-Carlo optimization problem \cite{mohamed2020monte} in DADA. The search of the DA policy and training of classification can be conducted simultaneously within this bi-level optimization framework. However, both categorical and Bernoulli distributions are not differentiable. To directly optimize policy hyper-parameters, it is necessary to perform gradient approximation on these non-differentiable prior to policy search. Inspired by DARTS \cite{liu2018darts}, DADA employs the Gumbel-Softmax gradient estimator \cite{jang2016categorical}, which is also known as a concrete distribution \cite{maddison2016concrete}. Such a gradient estimator is used to relax the distributions of augmentation operations. As for the operation hyper-parameters, i.e. operation probability and magnitude, an unbiased estimator RELAX \cite{grathwohl2017backpropagation} is applied to obtain their gradients with regard to model performance. 

The gradient relaxation in DADA consists of two parts, including the re-parameterization of categorical distribution for sub-policy selection, and the approximation of Bernoulli distributions for image operation hyper-parameters. In DADA, the sub-policy to augment data batches is selected from a categorical distribution. The preference for each sub-policy is defined using a probability parameter. After optimizing the parameter for categorical distribution, the sub-policies associated with higher probability will be selected to form the final DA policy. However, the parameter of sub-policy conforms to a non-differentiable distribution. Inspired by its success in the NAS field \cite{xie2018snas, dong2019searching}, DADA employs the Gumbel-Softmax estimator \cite{jang2016categorical} to approximate the gradient of parameters for sub-policy selection. For hyper-parameters to describe augmentation TFs, both application probability and magnitude are sampled from Bernoulli distributions. Similar to categorical distribution, Bernoulli distributions are not differentiable. To overcome the gradient issue, the same relaxation procedure is applied on Bernoulli distributions to obtain the gradient of TF hyper-parameters. Moreover, to mitigate the bias resulting from gradient approximations, DADA employs the RELAX estimator \cite{grathwohl2017backpropagation}, to achieve an unbiased gradient estimation, which further improves the policy search.

The major contribution of DADA is the innovation regarding gradient approximation. Instead of using a standard Gumbel-Softmax approximator, DADA applies unbiased RELAX estimator \cite{grathwohl2017backpropagation}, which estimates gradients more accurately. Through extensive experiments \cite{li2020dada}, DADA models using the RELAX gradient estimator achieve higher accuracy especially when compared with models using Gumbel-Softmax. DADA provides not only enhanced model performance, but also offers a significant speedup of the search process over alternative AutoDA approaches. Due to its increased efficiency, the search of DA policy in DADA is conducted on the full dataset instead of a reduced subset, which also improves the final results. In the field of automated DA policy search, the common sense is that using more data for the policy search will provide more information about the target task, and thus lead to a better final policy. On the other hand, a large amount of data will slow down the searching and raise time issues. This results in a trade-off between performance and efficiency in most AutoDA models. However, unlike prior methods such as AA \cite{cubuk2019autoaugment} and Fast AA \cite{lim2019fast}, DADA is able to well balance model accuracy and search costs in resource-constrained environment. Due to its efficiency, DADA is considered to be a feasible AutoDA approach for practical application.



\subsubsection{Automated Dataset Optimization (AutoDO)}

The majority of recent works in the AutoDA field focus on the reduction of the search cost, while Automated Dataset Optimization (AutoDO) \cite{gudovskiy2021autodo} evolves in the direction of mitigating the negative impacts of noisy or imbalanced data. To achieve a robust policy search, AutoDO adapts the idea of density matching \cite{lim2019fast} in a bi-level optimization framework. Specifically, the AutoDO model optimizes a set of augmentation hyper-parameters for each data point instead of a batch of data, allowing for more flexibility in tuning distributions of transformed data. In addition, AutoDO further refines the policy estimate by generalizing the training loss and softening the original labels. Through implicit differentiation, AutoDO jointly optimizes the results from three sub-models: the policy sub-model, the loss weighting sub-model and the soft label sub-model. Moreover, by using Fisher information \cite{domingos2020every, gudovskiy2020deep}, AutoDO provides theoretical proof that the complexity of AutoDA problem scales linearly with the size of the dataset. 

The proposal of AutoDO is mainly motivated by data problems present in training sets, including biased distributions and noisy labels. This issue becomes more predominant when existing approaches apply the same DA strategy on all data points for augmentation. For instance, data distributions of different classes are usually uneven in practice. However, by sharing the same augmentation policy among the entire training set, data samples in all categories are evenly augmented to increase diversity. Since the intensity of augmentation remains the same, classes with more data points might be over-augmented after transformation, while minority classes may be under-augmented. After data augmentation, an imbalanced or biased training set transformed by shared policy may potentially mislead the classification model. A shared DA policy is therefore not robust enough for data with distortions. For multi-class classification tasks, the overfitting issue may deteriorate significantly \cite{terhorst2021comprehensive}, especially when there exists noise in the original data labels \cite{zhang2021understanding}. This phenomenon is defined as the dilemma of shared-policy in \cite{gudovskiy2021autodo}. To overcome this limitation, AutoDO estimates DA hyper-parameters for each training data point, rather than the entire dataset. Additionally, the AutoDO algorithm considers loss weights to constrain distribution biases and soft labels to address label noises. 

A complete AutoDO model is composed of three sub-models: augmentation, loss re-weighting and soft-labelling sub-models. The overall workflow of AutoDO can be described as follows. Firstly, data are sampled as input to the augmentation sub-model, where the original data points are transformed by a set of point-wise augmentation operations. Each data sample is separately augmented by a sequence of specific transformation functions. The augmentation hyper-parameters for each image are defined and updated in the augmentation sub-model as well. To be more specific, application probabilities are binary values, while the magnitudes are sampled from a continuous Gaussian distribution. After data augmentation, the distorted data output from the augmentation block is used to train the classification network. During the training, the loss re-weighting sub-model and soft label sub-model are propagated at the same time. Specifically, the loss sub-model is used to normalize the training loss at certain training epoch, restraining the negative impacts of biased distributions. As for the soft label sub-model, it softens the original label of transformed data based on noise-free validation data. A soft labelling technique is applied in AutoDO to preclude potential noises oof data notation resulting from aggressive augmentation \cite{tanaka2018joint, yi2019probabilistic}. Lastly, the reward signal produced by loss re-weighting block, along with the soft labels from soft-labelling model are then back-propagated to update the augmentation hyper-parameters accordingly. 

In AutoDO, the optimization of the classification network and augmentation policy are conducted simultaneously. Inspired by prior works \cite{hataya2020faster, li2020dada, lin2019online}, AutoDO employs a bi-level setting, where the inner objective is to find the optimal network weights for target tasks, while the outer objective is to search for the optimal DA policy by hyper-parameter optimization. Such joint optimization is realized by gradient differentiation \cite{lorraine2020optimizing}. However, directly solving the bi-level problem is usually computationally infeasible, especially when AutoDO aims to optimize augmentation hyper-parameters per data point. To accommodate the search for large-scale hyper-parameters, AutoDO combines density matching techniques \cite{lim2019fast} to develop an implicit differentiation method. To be more specific, the major objective of searching in AutoDO is to minimize the distribution difference between the augmented data and an unbiased and clean validation set. According to analysis in \cite{gudovskiy2021autodo} and using the Fisher information, the modified differentiation framework can yield equivalent results to the DARTS gradient approximator \cite{liu2018darts} from previous methods \cite{hataya2020faster, li2020dada}. Furthermore, the use of Fisher information suggests a linear relationship between the complexity of AutoDA search and the size of task data. 

The effectiveness of AutoDO can be evaluated from two perspectives: class imbalance and label noise. Before being fed into AutoDO model, training data is distorted by adjusting the class distribution and associated labels. To display the strength of DA policy, t-SNE clustering method \cite{van2008visualizing} is employed to visualise the embedded features of test data in the classification model. The distance between data clusters represents the difference between data categories from the perspective of the model. Usually, larger margins or clear boundaries between clusters are preferable. When compared with gradient-based Fast AA \cite{lim2019fast}, the AutoDO model produces larger margins between data clusters in t-SNE plots. This result suggests that point-wise augmentation in AutoDO might achieve better performance. 

The extensive experiments in \cite{gudovskiy2021autodo} further confirms that AutoDO is more robust to distorted data. When compared with prior AutoDA approaches, AutoDO avoids overfitting to the majority data by optimizing point-wise hyper-parameters instead of a single shared policy. As a result, the augmentation policy learned by AutoDO can better separate images in different classes, mitigating the impacts of biased class distributions. The issue of noisy label is solved by using re-weighted loss and soft labels. From the experimental results, when trained by noisy data, AutoDO achieves superior results compared to other AutoDA models. Additionally, the smooth labels provided by the soft-labelling sub-model further enlarge the margins between data clusters in t-SNE plots, enhancing the generation ability of the model. More importantly, such improvements can be found in both well- and under-represented classes, aligning the accuracy of minority categories. The greatest advantage of AutoDO is its resilience to low-quality data. Overall, AutoDO is considered more applicable for real-world tasks with imperfect data.

\section{Discussion}
\label{Sec: discuss}

Following the categorization in Fig. \ref{Fig: taxonomy}, a wide range of AutoDA methods have been covered in this survey. However, automation of data augmentation is still a relatively new concept and has not been fully addressed. The development of AutoDA techniques are still in their infancy. Although AutoDA models have the potential to become an essential component of the standard deep learning pipelines, there are still a number of difficulties to overcome in the future. In this section, we provide a discussion focusing on the current challenges in existing AutoDA methods, as well as some directions we believe important for future work.

\subsection{Search Space Formulation}

The policy search space defined by AA \cite{cubuk2019autoaugment} has been widely accepted as the standard setting. Most of the later AutoDA works reuse the same formulation as the basis of their search model. However, without any modification, the parameterization in AA can result in an enormous search space. Even with a limited range of available TFs, the search process of AA still requires extensive computational resources, so it is not feasible in practice and has serious issues when scaling to larger datasets or models.

As AutoDA techniques evolve, traditional settings of the search space in AA-based models is challenged by search-free approaches. These methods aim to avoid the search phase by re-parameterization. For example, in RandAugment (RA) \cite{cubuk2020randaugment}, instead of optimizing the application probability, all TFs share the same global probability value. Moreover, based on empirical evidence, the magnitude is set to be a global variable for all transformations. The final search space in RA is controlled by only two hyper-parameters, the size of classification models and training sets. None of these requires significant computation to be optimized. 

UniformAugment (UA) \cite{lingchen2020uniformaugment} also re-formulates the search space. Instead of optimizing hyper-parameters, UA proposes the invariance hypothesis. Sampling any policies from an invariant policy space can retain the original label information for the majority of the augmented data. The authors of UA argue that if an augmentation space is approximately invariant, then optimizing the augmentation policy within that space is unnecessary. Therefore, a simple uniform sampling of the  invariant space is sufficient to effectively enhance the model performance, which completely eliminates the need for searching. 

The emerge of RA and UA has been revolutionary. The removal of the search stage raises questions on the necessity and optimality of searching in traditional AutoDA methods. Particularly, the hypothesis of invariant augmentation space has the potential to completely solve the current efficiency bottleneck of search-based methods. Hence, we believe a viable topic for future research is to discover a methodology for finding invariant augmentation spaces in certain domains. Furthermore, it is worth trying to explore other ways of defining augmentation policy searches for even more simplified search spaces. 

\subsection{Optimal Augmentation Transformations}

Though extensive efforts have been put into the search of the hyper-parameters that describe image transformation functions, less attention has been paid to the selection of TFs to be applied. Conventionally, the available image operations in AutoDA models are from the PIL Python library. Nearly all image transformation functions in PIL are considered in later search phases. In AA, two additional augmentation techniques, Cutout \cite{devries2017improved} and SamplePairing \cite{inoue2018data} are also used. 

In the majority of later works, similar selections are made for fair comparison purposes. A few of them remove some image operations from the search list \cite{tian2020improving, ho2019population} while others add several new augmentation technique into their model \cite{naghizadeh2020greedy, naghizadeh2021greedy}. The decision of image operations is made empirically, with little theoretical selection strategy. Though there is discussion around the different impact of each TF in terms of different datasets or sub-regions, no one has systematically investigate the optimal augmentation transformation functions for AutoDA to search. We argue that the optimization of available image operations in AutoDA with various data may be another interesting direction for researchers to explore. When searching for augmentation policies, future users might have the option to switch on or off a certain type of transformation in different application scenarios, so that the obtained augmentation policy can be more tailored to the given task.

\subsection{Unsupervised Evaluation}

The existing AutoDA methods extensively use supervised evaluation to determine which augmentation policies to use for training. Normally, the most generalized model is trained by applying the best augmentation policy. The optimal DA policy is supposed to provide the most enhancement in data variety and quantity while still retaining salient image features.
However, in practice, it can be difficult or impossible to obtain accurately labeled data, especially for sensitive tasks \cite{shin2011autoencoder}. This is a great challenge for existing AutoDA methods, and gives rise to the emergence of self-supervised AutoDA. This possibility is also discussed in \cite{wei2020circumventing}. So far, only a few of AutoDA models support semi-supervised learning, such as SelfAugment \cite{reed2021selfaugment}. However, with the rapid development in AutoDA field, this may be different in the future.

\subsection{Biased or Noisy Data}

When evaluating the effectiveness of AutoDA models, most existing works presume that the training data is clean and balanced. This happens with no doubt when using benchmark datasets such as CIFAR-10/100 \cite{krizhevsky2009learning} and ImageNet \cite{deng2009imagenet}. But in real-world scenarios, things could be totally different. It is an all too common case when training dataset is not only insufficient but extremely biased with label noise. Experimental results in \cite{gudovskiy2021autodo} show that the distorted training data with imbalanced distribution and noisy label can bring negative impacts to AutoDA models and eventually lead to overfitting problems. Additionally, findings in \cite{wei2020circumventing} also indicate that aggressive augmentation transformations might introduce label noise even though the original annotation is correct.

Dealing with biased or noisy data is a great challenge for AutoDA model. AA-KD \cite{wei2020circumventing} is the first AutoDA work targeting this issue. By leveraging the idea of KD, a stand-alone model is applied in AA-KD to provide extra guidance for model training. During the training, the model receives supervision from both ground-truth label and teacher signal, in case the discriminative information is accidentally removed by aggressive augmentation. AutoDO \cite{gudovskiy2021autodo} uses a similar idea to KD by softening the original label to better train the model. Additionally, AutoDO contains another re-weighting sub-model which is used to normalize the training loss. The combination of all three sub-models makes AutoDO much more robust than other AutoDA algorithms when given biased or noisy data. 

Overall, both methods outperform other AutoDA methods especially when dealing with imbalanced data with label noise. This further proves that AutoDA model can benefit from the extra handling of label noises. In the future, we expect that more research could tackle this topic. Such developments would greatly improve the applicability of AutoDA methods in real-world tasks.  

\subsection{Application Domains}

Though the major focus of this paper is on the image classification tasks, we identify several other domains which might greatly benefit from the AutoDA technique. One is Object Detection (OD) tasks, which has already been partly explored in recently published works, including AutoAugment for OD \cite{zoph2020learning}, DADA \cite{li2020dada}, RandAugment \cite{cubuk2020randaugment}, SelfAugment \cite{reed2021selfaugment} and Scale-aware AutoAugment (SA) \cite{chen2021scale}. Data augmentation may be even more important for OD tasks, especially when annotation is much more time-consuming. The experimental results in \cite{zoph2020learning} demonstrate that even a direct transfer of DA policies obtained from classification data can be useful for OD tasks. However, according to their findings, such improvement cmay be limited. Moreover, the extremely long searching time is also a serious problem in terms of applicability. 

To further improve the overall model performance, additional adjustment to the original AutoDA scheme must be done. Later works such as RA \cite{cubuk2020randaugment} concentrate more on the improvement in efficiency, which provides competitive accuracy with AA and superior search speed. Self-supervised evaluation is also found useful for detection training \cite{reed2021selfaugment}. DADA shows another possibility of tweaking the pre-trained backbone network used for detection, instead of directly operating on the detection model. The experimental results in \cite{li2020dada} indicate that pre-training the backbone network using DADA can improve the model performance for later detection tasks. Scale-aware Augmentation \cite{chen2021scale} is specifically designed for detection tasks by incorporating bounding-box level augmentations. Such design is more fine-grained and thus leads to considerable improvements to various OD networks.

While the aforementioned research mainly focuses on computer vision tasks, Natural Language Processing (NLP) is another field that can greatly benefit from the application of AutoDA techniques. Traditional data augmentation has been used extensively in NLP research. Automating the augmentation procedure in text-based tasks can be another promising research direction. Several studies already take a step on the application of AutoDA to linguistic problems, including the earliest TANDA \cite{ratner2017learning}, Text AutoAugment \cite{ren2021text} and works such as \cite{niu2019automatically, hu2019learning}. The adaptation of AutoDA methods in NLP yields competitive performance gains by improving the quantity and quality of training data.

\section{Conclusion}

With the increasing development of deep learning, training performant deep model efficiently largely depends on the quantity and quality of available training data. Data Augmentation (DA) is an essential tool for solving data problems, and been widely used in various computer vision tasks. However, designing an effective DA policy still highly relies on human efforts. It is difficult to select the optimal augmentation policy when given a specific dataset without domain knowledge. Therefore, researchers seek to solve this problem by automating the search of augmentation policies via deep learning, which stimulates the development of Automated Data Augmentation (AutoDA) techniques. 

This survey provides a comprehensive overview of AutoDA techniques for image classification tasks in the computer vision field. The focus of this paper is on various search algorithms in AutoDA. In order to describe and categorize approaches for augmentation policy optimization, we introduce searching and training phases for a standard AutoDA pipeline. Based on different optimization approaches, all AutoDA methods can be divided into two-stage or one-stage approaches. The searching process in AutoDA can be further classified into gradient-free, gradient-based or search-free methods. The associated qualitative evaluation describes AutoDA methods in terms of the complexity of the search space, the computational cost, the available augmentation transformations, as well as the reported performance improvements on classification models. In the future, we expect more research in the AutoDA field aimed at balancing between accuracy and efficiency of the proposed models, and providing better solutions to trade-off between safety and variety of the obtained augmentation policies.


\bibliography{sn-ref}


\end{document}